\RequirePackage{amsthm}
\documentclass[sn-mathphys,Numbered]{sn-jnl}
\usepackage{silence}
\usepackage{lmodern}
\usepackage{anyfontsize}
\usepackage{graphicx}%
\usepackage{multirow}%
\usepackage{amsmath,amssymb,amsfonts}%
\usepackage{amsthm}%
\usepackage{mathrsfs}%
\usepackage[title]{appendix}%
\usepackage{xcolor}%
\usepackage{textcomp}%
\usepackage{manyfoot}%
\usepackage{booktabs}%
\usepackage{algorithm}%
\usepackage{algorithmicx}%
\usepackage{algpseudocode}%
\usepackage{listings}%
\usepackage{subcaption}
\usepackage{float}

\theoremstyle{thmstyleone}%
\theoremstyle{thmstyletwo}%
\theoremstyle{thmstylethree}%

\begin{document}

\title[Multimodal Drowsiness Dataset]{UL-DD: A Multimodal Drowsiness Dataset Using Video, Biometric Signals, and Behavioral Data}

\author[1]{\fnm{Morteza} \sur{Bodaghi}}\email{bodaghi.morteza@gmail.com}

\author[1]{\fnm{Majid} \sur{Hosseini}}\email{mjhoseiny@gmail.com}
\author*[1]{\fnm{Raju} \sur{Gottumukkala}}\email{raju.gottumukkala@louisiana.edu}
\author[1]{\fnm{Ravi Teja} \sur{Bhupatiraju}}\email{ravi-teja.bhupatiraju@louisiana.edu}

\author[2]{\fnm{Iftikhar} \sur{Ahmad}}\email{iftikhar.ahmad@tietoevry.com}
\author[3]{\fnm{Moncef} \sur{Gabbouj}}\email{moncef.gabbouj@tuni.fi}

\affil*[1]{\orgname{University of Louisiana at Lafayette}, \country{USA}}

\affil[2]{\orgname{Tietoevry}, \country{Finland}}
\affil[3]{\orgname{Tampere University}, \country{Finland}}

\abstract{In this study, we present a comprehensive public dataset for driver drowsiness detection, integrating multimodal signals of facial, behavioral, and biometric indicators. Our dataset includes 3D facial video using a depth camera, IR camera footage, posterior videos, and biometric signals such as heart rate, electrodermal activity, blood oxygen saturation, skin temperature, and accelerometer data. This data set provides grip sensor data from the steering wheel and telemetry data from the American truck simulator game to provide more information about drivers' behavior while they are alert and drowsy. Drowsiness levels were self-reported every four minutes using the Karolinska Sleepiness Scale (KSS). The simulation environment consists of three monitor setups, and the driving condition is completely like a car. Data were collected from 19 subjects (15 M, 4 F) in two conditions: when they were fully alert and when they exhibited signs of sleepiness. Unlike other datasets, our multimodal dataset has a continuous duration of 40 minutes for each data collection session per subject, contributing to a total length of 1,400 minutes, and we recorded gradual changes in the driver state rather than discrete alert/drowsy labels. This study aims to create a comprehensive multimodal dataset of driver drowsiness that captures a wider range of physiological, behavioral, and driving-related signals. The dataset will be available upon request to the corresponding author.}


\maketitle

\section{Introduction}\label{sec1}
Driver drowsiness is an important public safety concern, contributing to thousands of fatalities annually and elevates crash risk. Early drowsiness detection can prevent dangerous micro-sleep episodes before they cause accidents. Numerous vehicular drowsiness detection systems have emerged to address this public safety concern, and beginning in 2024, inclusion of early drowsiness system is mandated in all new vehicles within the European Union. The performance of these systems is dependent on the detection approach. Developing robust drowsiness detection systems requires high-quality, publicly available datasets.

Traditional detection systems have relied heavily on video-based monitoring of facial cues, such as blink frequency, eye closure duration, and yawning patterns \cite{magana2017toward, weng2017driver, salzillo2020evaluation}. These techniques tend to be too late for timely intervention and are prone to degradation under poor lighting conditions or due to inter-individual variability \cite{hwang2016driver}. Physiological markers such as heart rate variability, skin temperature, and electroencephalogram (EEG) signals are alternative sensing technologies that enable early drowsiness detection \cite{massoz2016ulg, zheng2017multimodal}. However, these signals have subject-specific variability \cite{shen2021multi, esteves2021automotive}. A multimodal approach that integrates facial expressions, physiological responses, and subtle behavioral patterns, such as steering micro-corrections, offers the opportunity to develop more robust methods for driver drowsiness detection.

Despite the increased interest in multimodal approaches, most of the existing public datasets primarily focus on single-modality signals (e.g., video or biometrics). This limits their practical use in advanced drowsiness detection, where multimodal data is crucial. To address this, we collected a multimodal dataset named the University of Louisiana Drowsiness Dataset (UL-DD) from a lab study with 19 different subjects with facial video (RGB, IR, and 3D Depth camera), biometric signals (Electrodermal Activity (EDA), Blood Volume Pulse (BVP), Inter-Beat Interval (IBI), Saturation of Peripheral Oxygen (SPO2), Respiration Rate (RR), HR, and ST), and behavioral signals (driver telemetry, posture video, grip pressure sensors, and accelerometer (ACC) data) to provide an open dataset for researchers for early drowsiness detection. This multimodal approach improves drowsiness detection by leveraging complementary information from physiological, vision, and behavioral data. For example, facial cues of drowsiness appear when the person is already drowsy; however, a physiological signal like HR can provide small signs of fatigue earlier, which can be combined with other data types for richer information \cite{schleicher2008blinks}. 

We believe that this is a novel dataset that has the potential to allow researchers to develop metrics that may result in more early intervention by detecting latent and subtle declines in alertness for more timely interventions. Compared to other datasets, we collect a much more comprehensive set of streams from vision, biometry, and telemetry sensors.

The rest of the paper is organized as follows: Section \ref{sec2} reviews related work, comparing UL-DD to existing public datasets. Section \ref{sec3} describes the methodology, including participant recruitment, data collection protocols, and the collected multimodal signals. Section \ref{sec8} details the data records, file structures, and Synchronization. Section \ref{section5} provided future opportunities and guidelines for multimodal analysis. Section \ref{sec10} presents technical validation, including physiological and statistical analyses to confirm the dataset’s quality. Section \ref{sec14} provides usage notes, while Section \ref{sec15} outlines code availability. Finally, Section \ref{sec16} discusses limitations and future work, followed by acknowledgments and references.

\section{Related Work}\label{sec2}

\begin{table}[h]
\caption{Drivers' drowsiness detection publicly available datasets.}
\begin{tabular}{|l|l|l|l|}
\hline
Dataset            & Subjects                        & Signals                            & No of Classes                                                                                                    \\ \hline
RLDD               & 60 (51 M, 9 F)            & 30 Hours of RGB Video                                   & 3                                                           \\ \hline
NTHU-DDD           & 36                        & IR Video                                                                                                                                            & 5                  \\ \hline
DROZY              & 14 (3 M, 11 F)            & EEG, EOG, EKG, EMG, and NIR                               & 8                                                                                    \\ \hline
YawDD              & 107 (57 M, 50 F)          & Video   &                             4       \\ \hline
DMD                & 37                        & Video (capturing Face, Body, and Hand)                                                                                                                                                                                                                    & 3                  \\ \hline
3MDAD              & 50 (38 M, 12 F)                    & Video (RGB, IR, and Depth)                                                                                                                         & 2  \\ \hline
MRL Eye            & 37  (33 M, 4 F   & 15000 infrared images                                                                                                                & 2      \\ \hline
NITYMED            & 21 (11 M, 10 F) & Video                                                                                & 5                                                                                  \\ \hline
SEED-VIG           & 23                        & EEG and EOG signals (eye movement)                                                                                   & Score                                                                      \\ \hline
ULDD  & 19  (15 M, 4 F)                           & \begin{tabular}[c]{@{}l@{}}Video (RGB, IR, Pose), \\ Grip pressure, \\ Telemetry ( Heading, Pitch, Roll, Speed,\\ RPM, Gear), \\ Biometric Signals (HR, ST, ACC,\\ EDA, BVP, IBI, SpO2)\\ Facial Landmarks\\ Facial Action Units\\ Pose Landmarks\end{tabular}                                                                               & 9         \\ \hline
\end{tabular}
\label{tab:review}
\end{table}

Monitoring driver state has been an active area of research over the past few years. In this section, we will review the publicly available datasets, their collection methods, and the types of signals captured.
\cite{ghoddoosian2019realistic} collected 30 hours of RGB video from 60 subjects. Labels are alert, low vigilant, and drowsy. The video was captured by participants using their own mobile or web camera in different places. However, this dataset is limited to video signals with 10-minute durations for each subject, and it was not collected in a driving situation. NTHU-DDD \cite{weng2017driver} is a public dataset of IR videos from 36 people playing a driving simulation game in different situations, like normal driving, blinking slowly, yawning, falling asleep, and even laughing. However, it is based on subjects pretending to be drowsy, which might not be useful for detecting real drowsiness.  DROZY \cite{massoz2016ulg} is a multimodal dataset of EEG, EOG, ECG, EMG, and near-infrared images (NIR) captured from 14 participants in a lab environment using the same camera to capture different levels of drowsiness. This dataset is limited to 30 minutes of data for each user, and it is labeled every 10 minutes, which does not capture subtle changes in driver state. YawDD \cite{abtahi2014yawdd} includes videos recorded from two different angles in a parked car. 107 subjects participated in three situations: regular driving without talking, driving while talking or singing, and yawning while sitting in a parked car. NCKUDD \cite{chiou2019driver} dataset has videos of 25 participants filmed in a parked car in both daylight and dark conditions. The videos cover different situations like regular driving, feeling sleepy, being distracted, talking, eating, talking on the phone while driving, and other unusual driving behaviors.  SUST-DDD \cite{yilmaz2022sust} dataset consists of 2074 videos, each 10 seconds from 19 participants, recorded by drivers' phones during real driving, capturing moments of fatigue and normal driving. The dataset includes various lighting conditions and resolutions. It was labeled as drowsy or not drowsy based on watching the videos and voting. 3MDAD \cite{jegham2020novel} used two cameras from frontal and side views in a real driving car to collect data from 60 people. The dataset provides RGB, infrared, and depth frames for each view. The MRL Eye \cite{fusek2018pupil} collected a total of 15,000 infrared images of the human eye, which are captured in various lighting conditions. The images are labeled open or closed, with the presence of glasses, reflections, gender, and subject ID. The NITYMED dataset \cite{nuralif2023driver} dataset is a public collection of videos of drivers in real cars moving under nighttime conditions. It contains 130 videos with different features and levels of drowsiness. The SEED-VIG\cite{zheng2017multimodal} public dataset is a collection of EEG and EOG signals, as well as the corresponding vigilance level annotations, of subjects who played a driving game in a virtual driving system. The duration of the entire experiment was approximately 2 hours. DD-Pose \cite{roth2019dd} contains 330k measurements from multiple cameras acquired by an in-car setup during naturalistic drives. It also provides head pose annotations, steering wheel and vehicle motion information, and occlusion labels. DD-Pose is a diverse benchmark that can be used to develop and test head pose estimation methods for automotive applications. A vision-based public driver monitoring study\cite{ohn2013power} proposes a method for detecting and recognizing hand gestures and hand-object interactions using a camera pointed at a car dashboard.  The study uses RGB and depth images captured by a Kinect sensor, including 7207 sample frames that can be used for normal driving and distraction analysis.

Some studies collected their private dataset capturing features of the eye, mouth, and BPM signal \cite{choi2018driver}, video data of drivers in real driving road situations for eye tracking \cite{friedrichs2010camera}, RGB videos of participants in a parked car or simulator \cite{faraji2021drowsiness, {yang2024video}}, RGB, depth, and IR videos from 3 cameras capturing the face, body, and hands \cite{ortega2020dmd}, thermal, NIR, visual, and physiological data \cite{das2021multimodal}, driving performance, eye movement, and EEG \cite{guo2022driver}. Table \ref{tab:review} summarizes publicly available datasets, the number of subjects, collected signals, and class labels.

Comparing UL-DD to multimodal datasets in broader fatigue and sleep research, rather than solely driver-specific studies, shows its versatile usage and value. For instance, \cite{shi2023fatigue} provides EEG and eye-tracking data from 23 subjects and classifies 3-level fatigue states, but lacks video or diverse biometrics. \cite{kalanadhabhatta2021fatigueset} offers wearable sensor signals for mental fatigue detection without real-time behavioral or video data. \cite{stappen2021multimodal} presents a dataset with video and audio with a length of 40 hours for sentiment and fatigue analysis, yet misses depth video and physiological granularity. \cite{tao2024multimodal} focuses only on brain, muscle, and eye activity for driver behavior analysis missing actual behavioral signals like pose or telemetry data. 

Unlike these, UL-DD stands out with three key contributions: 1- continuous 40-minute sessions that capture small and gradual changes in drowsiness; 2- a unique combination of multimodal signals (video, biometric, behavioral), and 3- detailed 9-level annotations to offer more precision in driver state. This positions the UL-DD as a unique resource, not only for driver monitoring but also for advancing broader fatigue and sleep studies.

\section{Methods}\label{sec3}
\begin{figure}[t]
    \centering
    \includegraphics[width=0.5\linewidth]{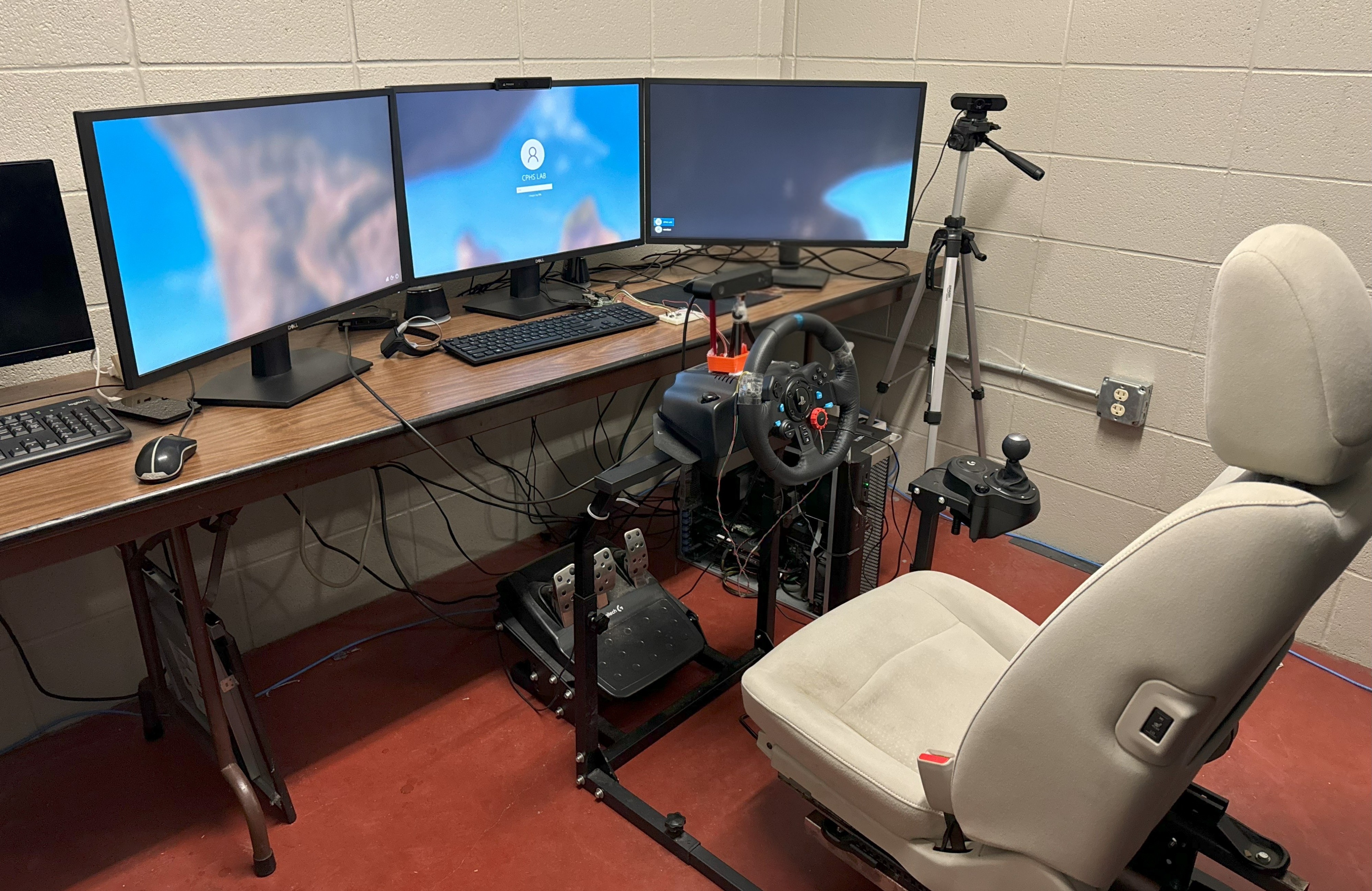}
    \caption{UL-DD driving simulator.}
    \label{fig:Simluator}
\end{figure}

We used American Truck Simulator as the driving simulator software. The software simulated trucking on U.S. highways, urban roads, and rural routes, with dynamic traffic and weather options. The simulator hardware included a three-monitor panoramic setup with a steering wheel and pedals. We did not use the manual gear system of the simulator. The subjects wore the following sensors: two wrist sensors and two on-the-wheel grip pressure sensors. They were observed by three cameras. The camera angles were depicted in Figure \ref{fig:Pose}. 19 subjects, each participated in two 40-minute simulator sessions for a total of 80 minutes. Figures \ref{fig:Simluator} and \ref{fig:devices} show the driving simulator and sensors, respectively.

\subsection{Participants}\label{sec4}
The participants were recruited primarily via flyers and recruitment emails in the engineering department of the University of Louisiana at Lafayette. While 30 subjects responded to the call, only 20 attended the study. Of these, 3 subjects only completed the first session and did not return to the second session. This left the study pool with 17 subjects who completed both sessions. We also included the 3 subjects who completed just the first session. One subject in the 2 session pool withdrew consent, leaving us with 16 subjects with 2 sessions and 3 subjects with 1 session, for a total of 19 subjects.

The age of the subject population ranged from 25 to 40 (M=27.3, STD=2.3). Other demographic information was not collected for privacy reasons. Inclusion criteria included individuals with driving experience, good general health, and no sleep disorders or cardiovascular conditions. The participants were asked to abstain from caffeine, tobacco, and other stimulants ten hours before the experiment. Further, participants were asked to avoid alcoholic beverages 24 hours before the study. Participation was voluntary and no incentives for participation were provided. Participants were informed that the cameras would collect their facial features and body posture during the tasks. The participants were also informed that the collected data would be released. The entire study protocol was supervised and approved by the UL Lafayette Institutional Review Board (IRB-22-038-IRI).

\subsection{Video data}\label{sec5}
\begin{figure}[h]
    \centering
    \begin{subfigure}[b]{0.3\textwidth}
        \centering
        \includegraphics[width=\linewidth]{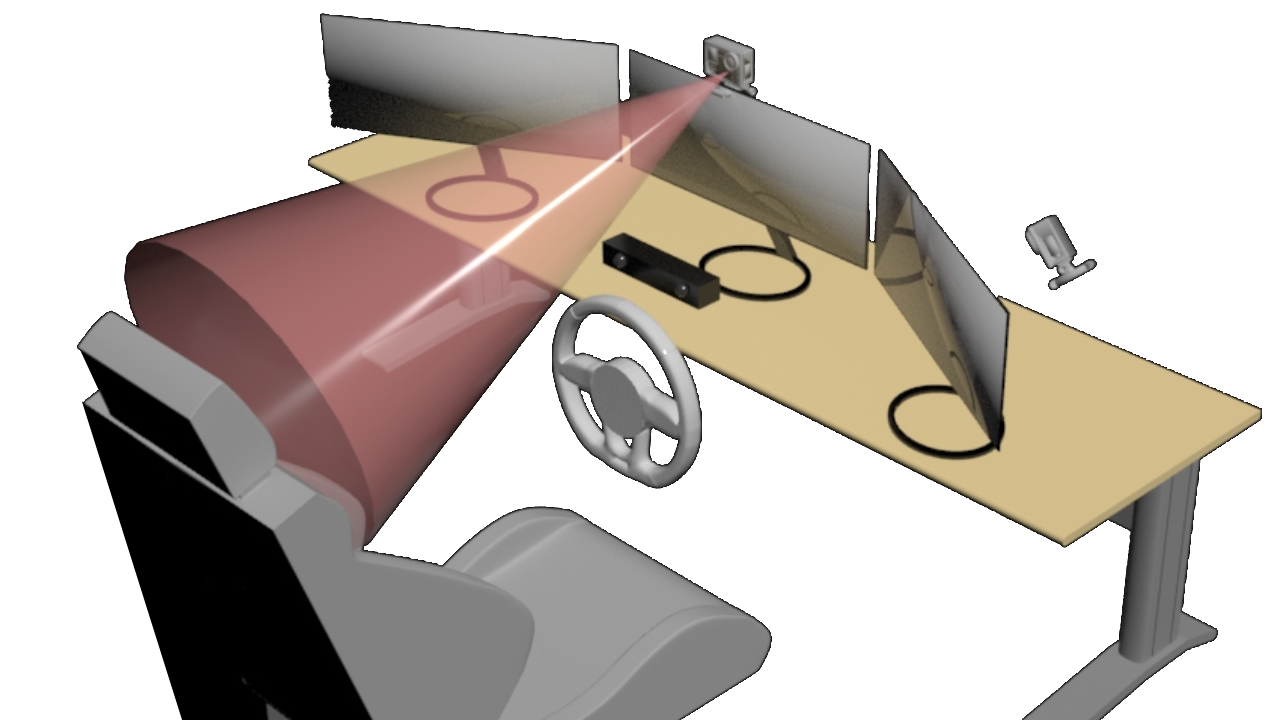}
        \caption{Direction of the infrared camera.}
        \label{fig:IR}
    \end{subfigure}
    \hfill
    \begin{subfigure}[b]{0.3\textwidth}
        \centering
        \includegraphics[width=\linewidth]{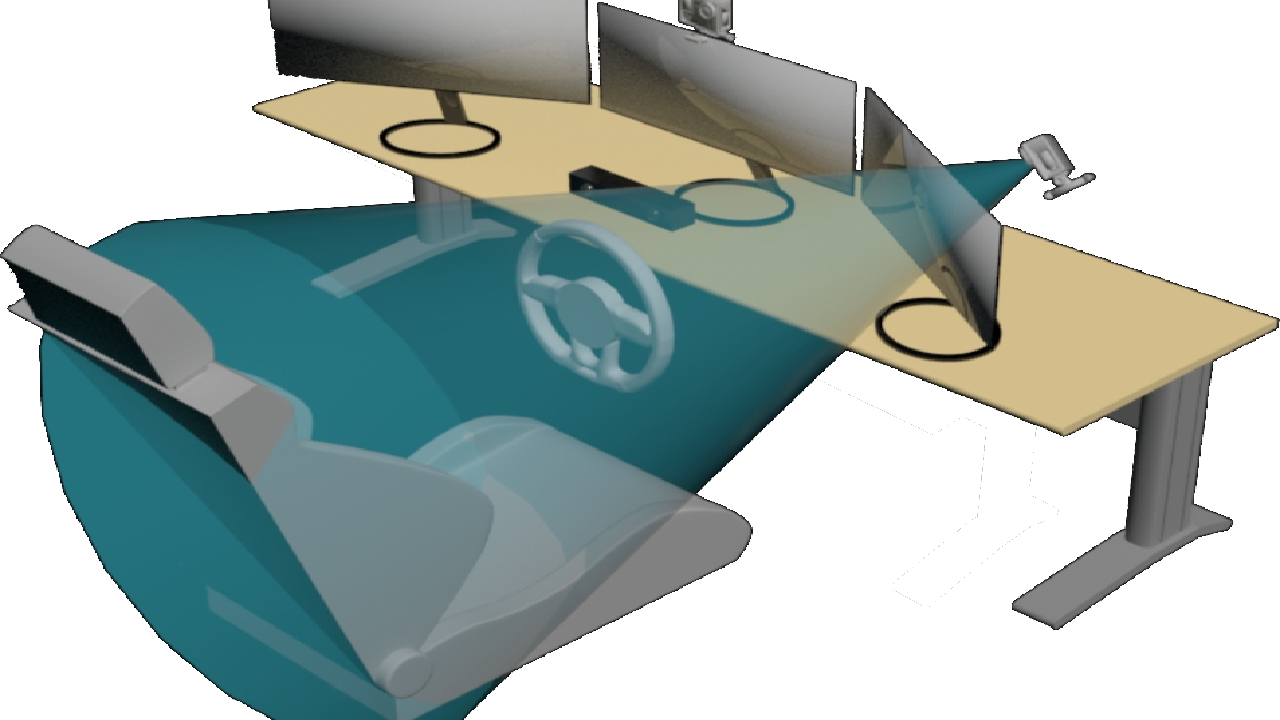}
        \caption{Direction of the pose camera.}
        \label{fig:subfig2}
    \end{subfigure}
    \hfill
    \begin{subfigure}[b]{0.3\textwidth}
        \centering
        \includegraphics[width=\linewidth]{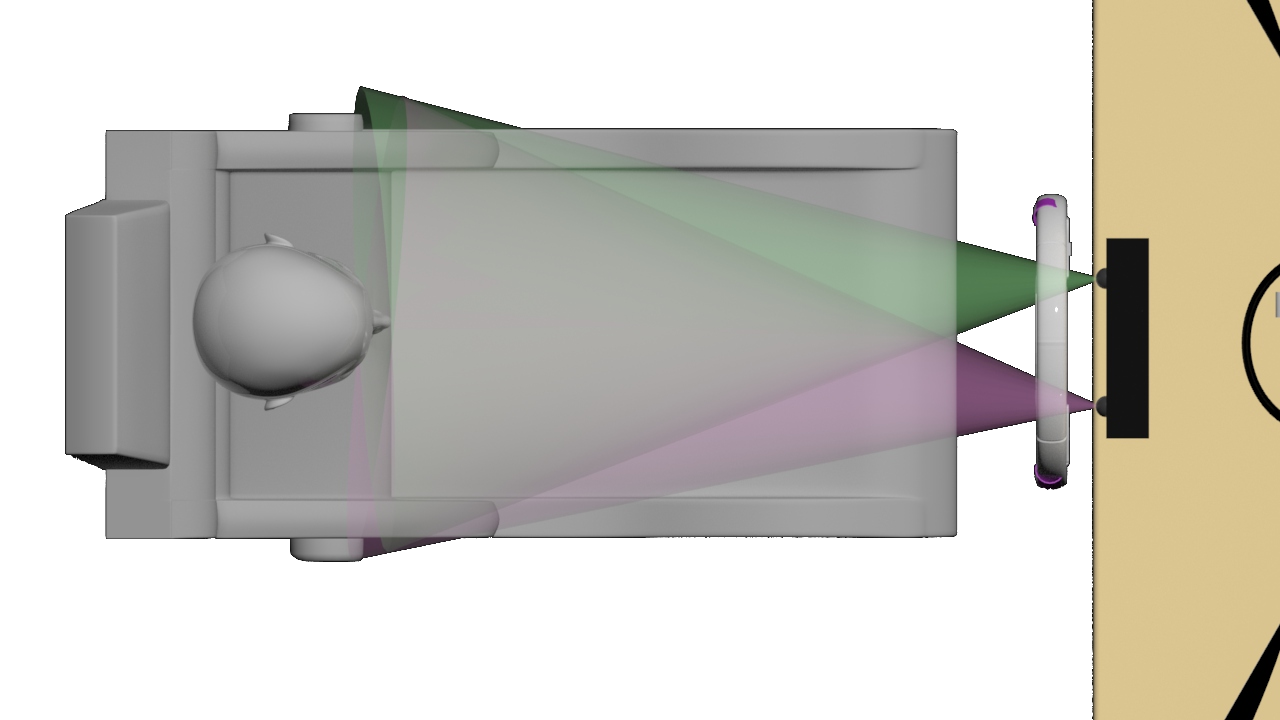}
        \caption{Direction of the Zed 2 3D camera.}
        \label{fig:subfig3}
    \end{subfigure}
    
    \caption{Position and direction of different cameras in the simulator.}
    \label{fig:Pose}
\end{figure}

We used three different cameras during data collection to record the driver's face and movement. Figure \ref{fig:Pose} shows the direction of all three cameras in the simulator setup.

\subsubsection{ZED 2 Depth Camera}
In this data collection, a "Zed 2" 3D depth camera (Figure \ref{fig:zed2}) was used for a special perception and environmental understanding. For each subject, the recorded video size is approximately 40 GB, with a resolution of 1344x376 at 60 frames per second, saved in MP4 format. The 3D camera was placed behind the steering wheel and next to the monitors to capture depth information without letting the steering wheel block the recording of facial data. We also shifted the camera and monitor positions to make sure the 3D camera did not get in the way of the drivers seeing the screens clearly. Figure \ref{fig:subfig3} illustrates the position and direction of the 3D camera.  For public access, the video was split into two separate views (left and right), each resized to 440x370 resolution, which greatly reduced the file size and made it simpler to work with.

\subsubsection{Infra-Red (IR) Camera}
A 1080p IR camera was used to capture facial video, shown in Figure \ref{fig:IRcamera}. This camera has a high resolution and the ability to perform well under various lighting conditions, especially in low light conditions, which is common in actual driving situations at night. For each participant, the recorded video file size is approximately 35 GB, with a resolution of 640x360 at 60 frames per second, saved in MP4 format. Figure \ref{fig:IR} shows the position and direction of the infrared camera. To focus on the facial region and remove the background, the video data was later cropped to a resolution of 240x220, which significantly reduced the file size to improve storage efficiency and accessibility.

\subsubsection{RGB Posture Camera}\label{sec3.2.3}
\begin{figure}[t]
    \centering
    \begin{subfigure}[b]{0.30\textwidth}
        \centering
        \includegraphics[width=\linewidth]{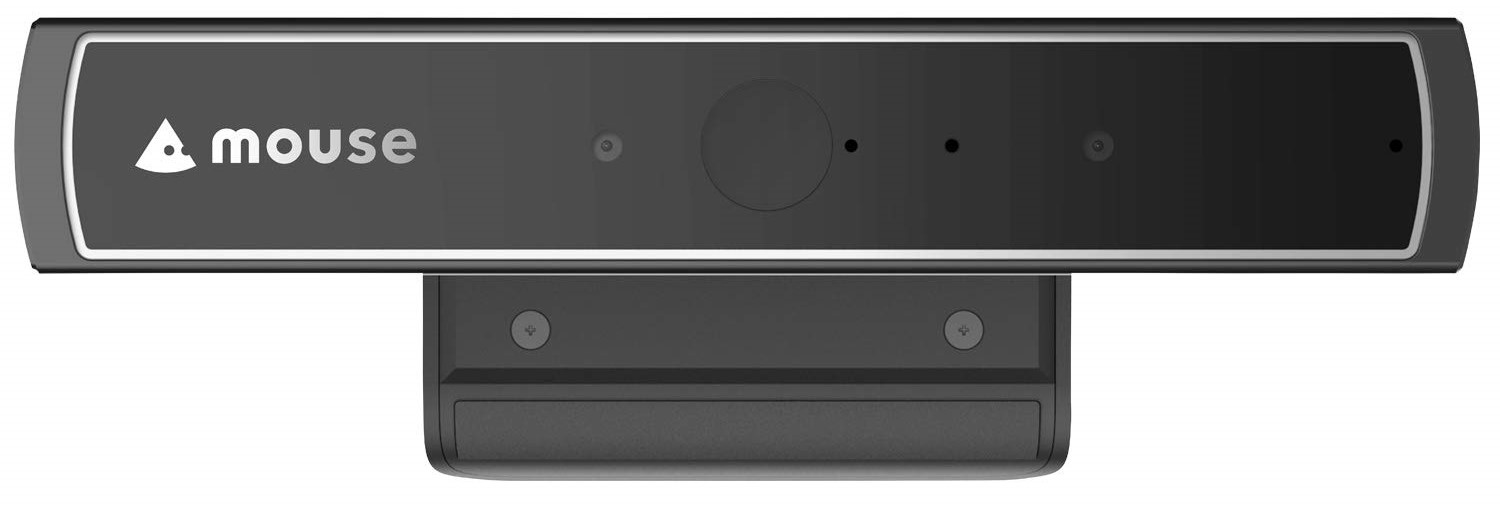}
        \caption{IR camera.}
        \label{fig:IRcamera}
    \end{subfigure}
    \hspace{-1mm}
    \begin{subfigure}[b]{0.40\textwidth}
        \centering
        \includegraphics[width=\linewidth]{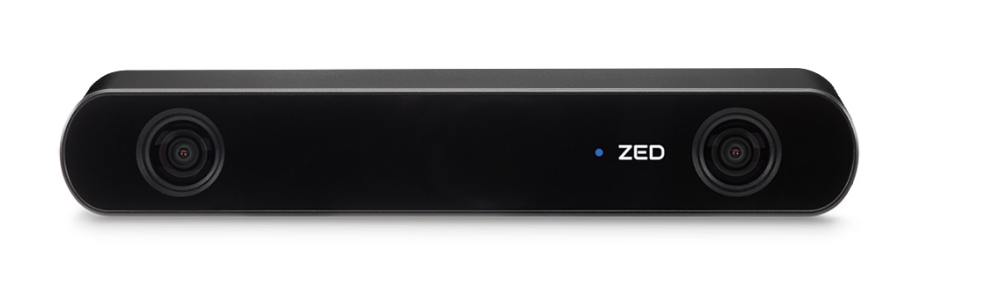}
        \caption{ZED 2 camera.}
        \label{fig:zed2}
    \end{subfigure}
    \hspace{-1mm}
    \begin{subfigure}[b]{0.28\textwidth}
        \centering
        \includegraphics[width=\linewidth]{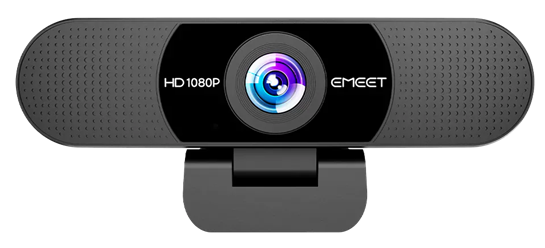}
        \caption{Pose camera.}
        \label{fig:posecamera}      
    \end{subfigure}
    \hspace{2mm}
    \begin{subfigure}[b]{0.17\textwidth}
        \centering
        \includegraphics[width=\linewidth]{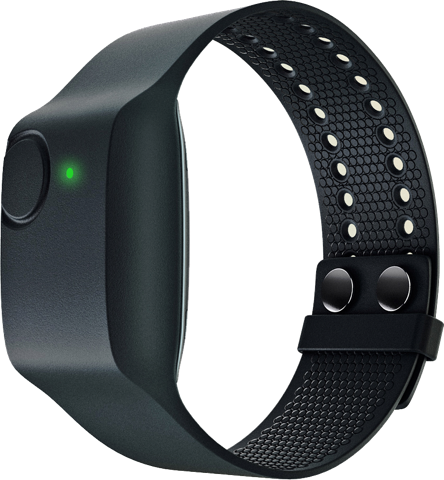}
        \captionsetup{justification=centering}
        \caption{Empathica E4 wristband.}
        \label{fig:E4}      
    \end{subfigure}
    \hspace{5mm}
    \begin{subfigure}[b]{0.22\textwidth}
        \centering
        \includegraphics[width=\linewidth]{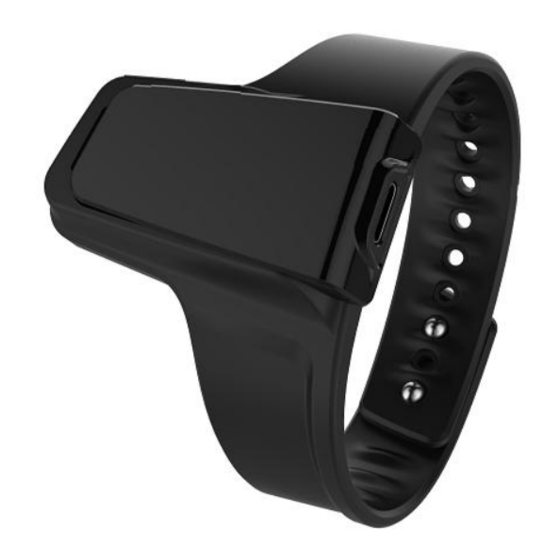}
        \captionsetup{justification=centering}
                \caption{O2 Max \\wristband.}
        \label{fig:O2}      
    \end{subfigure}
   \hspace{5mm}
    \begin{subfigure}[b]{0.13\textwidth}
        \centering
        \includegraphics[width=\linewidth]{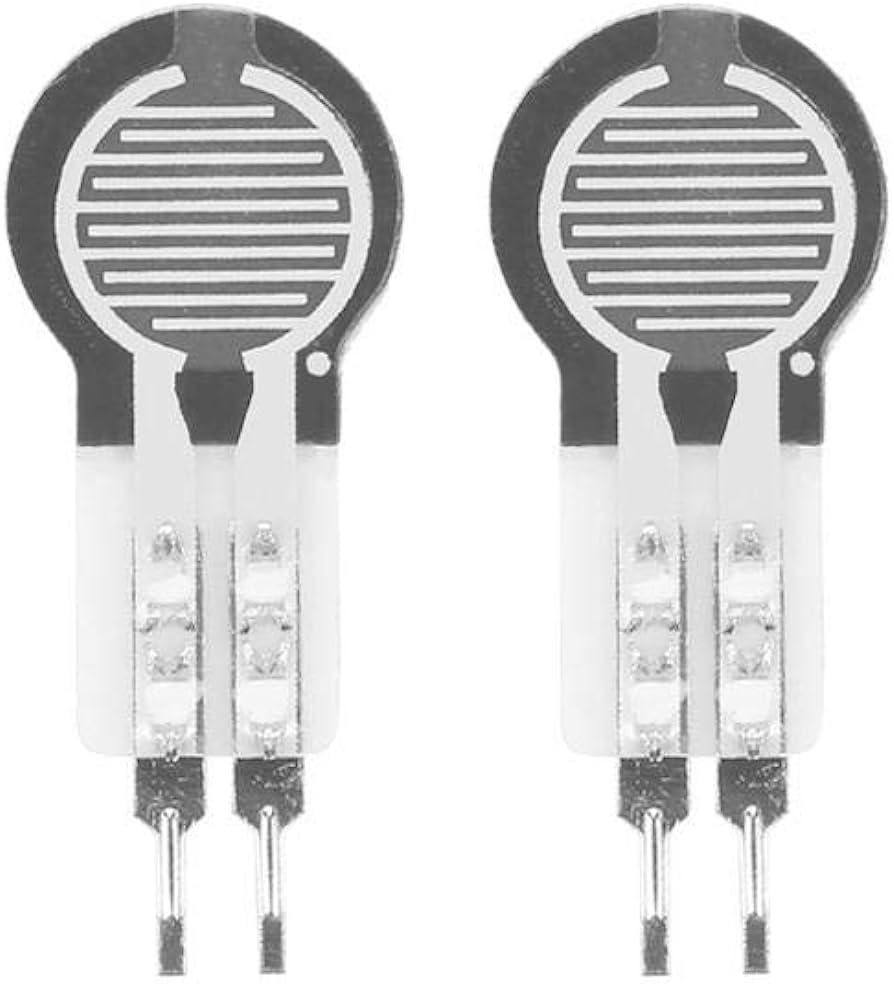}
        \captionsetup{justification=centering}
        \caption{Pressure Sensor.}
        \label{fig:PS}      
    \end{subfigure}
    \caption{Devices used for data collection.}
    \label{fig:devices}
\end{figure}

An EMEET Basic FHD 1080P Webcam (Figure \ref{fig:posecamera}) was used to 
capture posture data through RGB video to analyze the subjects' physical movement. This camera provides full HD resolution, giving sharp and reliable images of body posture. For each participant, the video file is about 38 GB in size, recorded at a resolution of 640x480 and 60 frames per second, and saved as an MP4 file. The position and direction of the posture camera are shown in Figure  \ref{fig:subfig2}. To cut out the background and save space, we trimmed the video to a resolution of 620x470, reducing the overall file size.

\subsection{Facial Features}
\label{FF}
\begin{figure}[H]
    \centering
    \includegraphics[width={0.7\textwidth}]{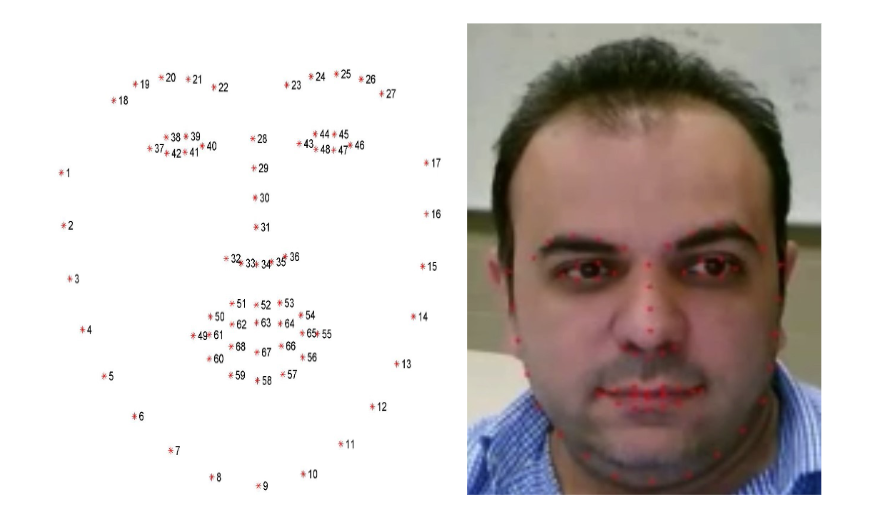}
    \caption{Participants’ 68 facial landmarks; this manuscript does not contain any images of the participants. We used one of the author's images to show the 68 facial landmarks.}
    \label{fig:68}
\end{figure}

We extracted 68 facial landmarks (FL) from the IR video for each user individually. This included subjects who did not consent to publication of video but consented to publication of features. All subjects consented to publication of facial features. Additionally, Facial Action Units (FAUs) were derived from these landmarks to capture key facial expressions and movements. Example FL points are shown in Figure \ref{fig:68}.

\begin{table}[htbp]
\caption{Facial Action Units.}
\begin{tabular}{|l|l|l|l|l|l|}
\hline
\textbf{No} & \textbf{Action Unit} & \textbf{Name}              & \textbf{No} & \textbf{Action Unit} & \textbf{Name} \\ \hline
1           & AU1                  & Inner Brow Raiser          & 16          & AU18                 & Lip Puckerer  \\ \hline
2           & AU2                  & Outer Brow Raiser          & 17          & AU20                 & Lip Stretcher \\ \hline
3           & AU4                  & Brow Lowerer               & 18          & AU22                 & Lip Funneler  \\ \hline
4           & AU5                  & Upper Lid Raiser           & 19          & AU23                 & Lip Tightener \\ \hline
5           & AU6                  & Cheek Raiser               & 20          & AU24                 & Lip Pressor   \\ \hline
6           & AU7                  & Lid Tightener              & 21          & AU25                 & Lips Part     \\ \hline
7           & AU9                  & Nose Wrinkler              & 22          & AU26                 & Jaw Drop      \\ \hline
8           & AU10                 & Upper Lip Raiser           & 23          & AU27                 & Mouth stretch \\ \hline
9           & AU11                 & Nasolabial Furrow Deepener & 24          & AU28                 & Lip Suck      \\ \hline
10          & AU12                 & Lip Corner Puller          & 25          & AU41                 & Lid droop     \\ \hline
11          & AU13                 & Cheek Puffer               & 26          & AU42                 & Slit          \\ \hline
12          & AU14                 & Dimpler                    & 27          & AU43                 & Eyes Closed   \\ \hline
13          & AU15                 & Lip Corner Depressor       & 28          & AU44                 & Squint        \\ \hline
14          & AU16                 & Lower Lip Depressor        & 29          & AU45                 & Blink         \\ \hline
15          & AU17                 & Chin Raiser                & 30          & AU46                 & Wink          \\ \hline
\end{tabular}
\label{tab:fau}
\end{table}

Facial Action Coding System (FACS) is a comprehensive framework for understanding facial expressions, particularly the discrete movements and gestures referred to as Action Units (AUs). Recent studies showed that using FAUs can reduce the amount of unwanted information from the video data, reduce the data size and computational cost, and increase the model performance by removing redundant information (noise) like skin tone, lighting, and background \cite{clark2020facial}. We generated and provided a set of action units along with the other data types, as listed in Table \ref{tab:fau}. However, we mostly focused on upper-face facial action units and yawing.

\subsection{Biometric Signals}\label{sec6}
We used two wristbands, the Empatica E4 (Figure \ref{fig:E4}) and Checkme O2 Max (Figure \ref{fig:O2}), for collecting biometric signals. The following biometric signals are provided in the dataset:
\subsubsection{Multi-sensor Wearable}
Below are the signals that were collected using Empatica E4:
\begin{itemize}
    \item \textbf{HR:} Heart Rate typically ranges from 60 to 100 beats per minute when a person is resting. However, this rate can change depending on environmental factors like physical activity, mood, and health condition. Wearable sensors track these changes and patterns, which may potentially offer insights about drowsiness states\cite{vicente2016drowsiness}.
    \item \textbf{TEMP:} Skin Temperature, another signal from the E4, serves as an indicator of peripheral vasodilation and has been associated with fluctuations in alertness and arousal levels\cite{raymann2007skin}. Wearable sensors capture small body temperature changes, which usually happen during sleep \cite{dolson2022wearable}. This can be studied with other physiological signals to detect signs of fatigue and drowsiness \cite{chowdhury2018sensor}.
    \item \textbf{EDA:} Also known as Galvanic Skin Response (GSR), measures sweat gland activity and gives information about emotional arousal and stress. Fluctuations in EDA can show physiological states that are related to drowsiness, such as low alertness and fatigue \cite{jiao2023driver}.
    \item \textbf{ACC:} Accelerometer data, measured by sensors, records object acceleration. These sensors have diverse applications, such as human action recognition and step counting. Wearable ACCs track movement patterns, offering information about activity levels\cite{lara2012survey}.
    \item  \textbf{BVP:} Blood Volume Pulse measures blood flow changes to assess cardiovascular activity. Variations in BVP can indicate physiological changes related to drowsiness \cite{saleem2023systematic}.
    \item \textbf{IBI:} InterBeat Interval represents the time between heartbeats. Changes in IBI patterns can reflect the autonomic nervous system and heart activity that may help detect early signs of decreased vigilance \cite{forcolin2018comparison}.
\end{itemize}

\subsubsection{Pulse Oximeter}
Below are the signals that were collected using Checkme O2 Max Pulse Oximeter:

\begin{itemize}
    \item \textbf{Pulse Rate:} It measures the number of heartbeats per minute and is identical to HR. However, since it is measured using different devices, we refer to them separately to avoid confusion. 
    \item \textbf{Motion:} Motion data, collected alongside SpO2, to help identify and flag potential motion artifacts in the SpO2 readings \cite{clarke2014effects}.
    \item \textbf{SpO2:} Saturation of peripheral oxygen measures blood oxygen levels and helps evaluate overall physical health. Drops or changes in SpO2 can be tied to uneven breathing patterns that happen during drowsiness, making it a helpful sign for drowsiness monitoring \cite{thomson2005ventilation}.
\end{itemize}

\subsection{Behavioral Data}
Behavioral data provides complementary information for biometric signals and facial data to detect drowsiness. We collect behavioral data to monitor cues such as posture, head tilting, grip pressure changes, and different driving patterns, which can be fatigue indicators \cite{dong2010driver}. We collected three types of behavioral data:

\begin{itemize}
    \item \textbf{Pose:} We captured posture data using a webcam, as explained in \ref{sec3.2.3}. We extracted pose landmarks (PL) from the video using MediaPipe Pose \cite{lugaresi2019mediapipe}, a real-time pose estimation framework. Pose changes such as slouching, head tilting, or leaning to one side may show fatigue or loss of focus during long driving sessions \cite{mahomed2024driver}.
    \item \textbf{Grip pressure:} Grip pressure data was collected using a pressure sensor installed on both sides of the steering wheel, shown in Figure \ref{fig:PS}. Changes in grip strength or asymmetry in pressure distribution can indicate drowsiness-related muscle relaxation or changes in driver engagement. Consistent pressure patterns may reflect alertness, while drops or fluctuations could correspond to drowsiness \cite{lee2016detecting}.
    \item \textbf{Telemetry:} Telemetry data was collected from the driving simulator software and offers more insights into driving patterns, such as speed and lane changing. Inconsistent speed or delayed reactions may indicate drowsiness \cite{baulk2001driver}.
\end{itemize}

\subsection{Data collection protocol}\label{sec7}
Our data collection took place in a controlled laboratory setting, where we used the American Truck Simulator software deployed in a custom-built driving simulator. We used three cameras for recording facial and pose data (see Section \ref{sec5}). We used two different wristbands to collect biometric signals (see Section \ref{sec6}). Additionally, other behavioral data like telemetry and grip pressure were collected through the game and sensors installed on the steering wheel, respectively.

\begin{table}[htbp]
 \caption{Description of the drowsiness data collection tasks and durations.}
\begin{tabular}{|llp{7.5cm}|l|}
\hline
\multicolumn{1}{|l|}{\textbf{No.}} & \multicolumn{1}{l|}{\textbf{Task}}          & \textbf{Action performed  }                                                                         & \textbf{Time}   \\ \hline
\multicolumn{1}{|l|}{\textbf{1}}    & \multicolumn{1}{l|}{Preparation}   & Familiarize the driver with the simulator & 10 min \\ \hline
\multicolumn{1}{|l|}{\textbf{2}}    & \multicolumn{1}{l|}{Rest}          & Sitting behind the wheel and relax before data collection                                          & 5 min  \\ \hline
\multicolumn{1}{|l|}{\textbf{3}}    & \multicolumn{1}{l|}{Driving} & Driving in a simulated space and staring data collection                 & 40 min \\ \hline
\multicolumn{1}{|l|}{\textbf{4}}    & \multicolumn{1}{l|}{Survey}        & Complete KSS questionnaire                                                                 & 5 min  \\ \hline
\multicolumn{3}{|l|}{\textbf{Total:}}                                                                                                                                & 60 min \\ \hline
\end{tabular}
\label{tab:tasks}
\end{table}

The data was collected from 19 subjects (15 males, 4 females) performing driving tasks shown in Table \ref{tab:tasks} at different drowsiness levels. We asked them to attend the driving sessions once they were alert and showed no signs of drowsiness, and once when they felt drowsy. Drowsiness was verified using the Karolinska Sleepiness Scale questionnaire, requiring a score of $\geq$6 (some signs of sleepiness) to start. For each session, the aim was to complete an assigned route; much of the route was on a simulated major limited highway. At first, we asked them to drive for 10 minutes to get familiar with the simulator, followed by a 5-minute rest, and then they drove for 40 minutes, approximately 40 to 50 virtual miles, depending on speed and traffic conditions. Participants followed the simulator’s GPS navigation to avoid other distractions beyond speed control and response to traffic. During driving, they were asked to report their drowsiness level based on the KSS questionnaires every four minutes. The labeled data of the active driving session is made available in the dataset.

\subsection{Drowsiness detection surveys}\label{sec:surveys}

Each task was divided into ten four-minute sub-intervals, totaling 40 minutes for each session. The KSS was chosen because it has a standard 9-level scoring to capture gradual changes in drowsiness, which is widely used for driving studies \cite{aakerstedt1990subjective}. Table \ref{tab:kss} shows the KSS questionnaires that were asked to evaluate participants' drowsiness levels from 1 to 9 for each sub-interval. To ensure the reliability of these scores, we calculated Cronbach’s alpha, a standard measure of reliability \cite{tavakol2011making}. The resulting $\alpha$ coefficient was 0.99, indicating excellent internal consistency. This means the repeated scores reliably captured how each participant’s drowsiness changed over time.

\begin{table}[htbp]
\caption{Karolinska sleepiness scale.}
\begin{tabular}{|l|l|}
\hline
\textbf{Score} & \textbf{Description}                                             \\ \hline
\textbf{1}    & Extremely alert                                         \\ \hline
\textbf{2}    & Very alert                                              \\ \hline
\textbf{3}    & Alert                                                   \\ \hline
\textbf{4}    & Rather alert                                            \\ \hline
\textbf{5}    & Neither alert nor sleepy                                \\ \hline
\textbf{6}    & Some signs of sleepiness                                \\ \hline
\textbf{7}    & Sleepy, but no effort to keep awake                     \\ \hline
\textbf{8}    & Sleepy, some effort to keep awake                       \\ \hline
\textbf{9}    & Very sleepy, great effort to keep awake, fighting sleep \\ \hline

\end{tabular}
\label{tab:kss}
\end{table}

\section{Data records}\label{sec8}

The UL-DD dataset includes three main folders of Video\_Data, CSV\_Files, and Extracted\_Features, along with a separate Labels.CSV file. Each Folder has 19 folders from A to S, which show the participant's ID. 
For each participant, there are two folders, A and D, which stand for the Awake and Drowsy sessions, respectively.
File names begin with the participant’s ID, followed by the signal name, and the session. For instance, $A\_IR\_D$ shows IR video data for participant A, Drowsy session, whereas $B\_FL\_A$ refers to FLs extracted from video data for participant B and awake session. The labels are also available for each participant separately. We labeled the data every four minutes, and the label files contained ten instances. A detailed description of the files is included in the following section.

\subsection{Data files description}\label{sec9}
The following is the description of files in the task sub-folders: 

\begin{itemize}
    \item \textbf{User\_HR\_Session.csv:} The file contains a single column that corresponds to the average heart rate. The HR signal was recorded at a frequency of 1 Hz.
    
    \item \textbf{User\_ACC\_Session.csv:} The file contains three columns corresponding to the x-axis, y-axis, and z-axis accelerometer data, respectively. The 3-axis ACC data was recorded at a frequency of 32 Hz.
    
    \item \textbf{User\_EDA\_Session.csv:} The file contains a single column that includes average electrodermal activity. The EDA signal was recorded at a frequency of 4 Hz.
    
    \item \textbf{User\_TEMP\_Session.csv:} The file is a single-column CSV format that contains the average skin temperature. The TEMP signal was recorded at a frequency of 4 Hz.
    
    \item \textbf{User\_BVP\_Session.csv:} The file is a single-column CSV format that contains the blood volume pulse signal. The TEMP signal was recorded at a frequency of 4 Hz.
    
    \item \textbf{User\_IBI\_Session.csv:} The file contains two columns: the first column shows the time stamp, and the second column contains interbeat interval values, which are the difference between two consecutive rows of the first column. This signal contains time, which means IBI can not be interpreted in Hz frequency.
    
    \item \textbf{User\_O2M\_Session.csv:} The file contains three columns: blood oxygen level, pulse rate, and motion activity data. The SPO2 signal was recorded at a frequency of 0.5 Hz.

    \item \textbf{User\_RGP\_Session.csv \& User\_LGB\_Session.csv :} Each file contains a single column that includes the grip pressure data on the steering wheel for the right and left hands. The GP data was recorded at a frequency of 3 Hz.

    \item \textbf{User\_Telemetry\_Session.csv :} The file contains ten columns: Timestamp, raw rendering timestamp, raw simulation timestamp, raw paused simulation timestamp, pitch, roll, speed, rpm, and gear. This data captures the driving behavior from the American Truck Simulator. The telemetry signals were recorded at a frequency of 60 Hz.
    
    \item \textbf{User\_FL\_Session.csv:} The file contains $68 \times 2 = 136$ columns representing the x and y coordinates of the participant's 2D FLs. These 68 landmarks were extracted from the "IR.mp4" video using the Dlib \cite{king2009dlib} library. The data was sampled at a frequency of 60 fps, with the corresponding frame number recorded in the first column.

    \item \textbf{User\_PL\_Session.csv:} The file contains $33 \times 3 = 99$ columns that present the x, y, and z coordinates of the participant's PLs. These Pls were extracted from the "Pose.mp4" video files using MediaPipe Pose\cite{lugaresi2019mediapipe}. The data was sampled at a frequency of 60 fps, with the corresponding frame number recorded in the first column.
    
    \item \textbf{User\_FAU\_Session.csv:} The file contains thirty columns that present the main thirty facial action units explained in section \ref{FF}. FAUs were extracted from the "FL.csv" files. The data was sampled at a frequency of 60 fps, with the corresponding frame number recorded in the first column.
    
    \item \textbf{User\_IR\_Session.mp4:} The video file contains raw trimmed infrared video files of participants who consented to release their videos. This file is not available for all participants. The video has a frequency of 60 fps.
    
    \item \textbf{User\_R3D\_Session.mp4 \& User\_L3D\_Session.mp4:}  These video files contain the right and left angles of raw trimmed depth camera recordings for participants who consented to release their videos. These files are not available for all participants, but the corresponding FLs are provided for all the subjects. The video was recorded at 60 fps.

    \item \textbf{User\_Pose\_Session.mp4:} This raw, trimmed video files capture a side view of the driver for participants who consented to release their videos. It is not available for all participants, but corresponding PLs are provided for all the subjects. The video was recorded at 60 fps.

    \item \textbf{Labels.csv:} The file contains ten columns, each representing the driver's drowsiness level assessed over a four-minute interval with the corresponding User\_Session information in the first column.
    
\end{itemize}

\subsection{Synchronization}
\label{sec:sych}

In multimodal learning, aligning data over time is essential to properly combine different data streams for successful model training and analysis. Data often comes from various sources, each with its own recording frequency, delay, and time reference, which can cause misalignment if not appropriately handled. Without good synchronization, timing differences can lead to mistakes, weaken model accuracy, and distort how modalities relate to each other \cite{iwama2024two}. There are several approaches to synchronization, depending on whether a shared reference is available. The best scenario would be using a single clock for all devices, but this isn’t always practical. A more feasible approach is timestamping, where each data point gets an exact time label, allowing it to line up data based on when it was actually recorded. This method ensures temporal consistency across different modalities, even when their sampling rates vary \cite{dolmans2020data}.

For our dataset, we synchronized all modalities using manually recorded session start and end times as reference points. Video data captured simultaneously from three cameras was trimmed to match the session window. Biometric signals from the E4 wristband were also synchronized using epoch timestamps, ensuring alignment with the session timeline. Similarly, physiological data collected from the Checkme O2 Max wristband, including SPO2, pulse rate, and motion,  contained actual clock timestamps, which facilitated direct alignment with the session times. Grip pressure data was synchronized by matching the recorded start time in its file to the manually recorded session start. Telemetry data, which began when the driving simulation started, was aligned by observing its initiation in the video recordings and trimming excess data accordingly.

In cases where interruptions occurred during data collection, such as software crashes, we relied on notes taken during the sessions and video recordings to identify and correct the affected parts. In this way, all modalities remained consistently aligned despite these interruptions.

\section{Future Opportunities}\label{section5}

\subsection{Multimodal Analysis Guidelines}\label{pre}

\begin{figure}[htbp]
    \centering
    \includegraphics[width={1\textwidth}]{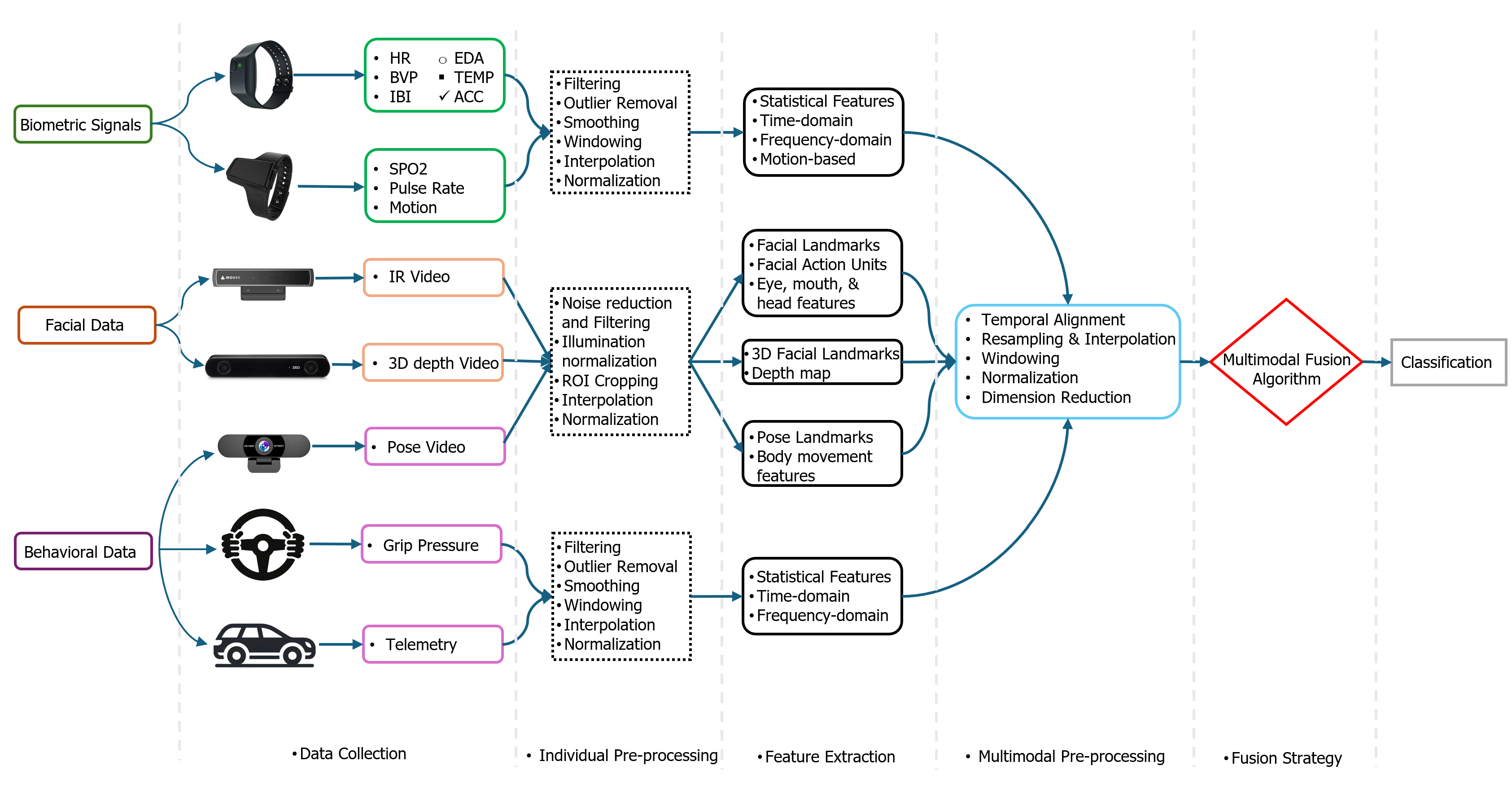}
    \caption{Multimodal analysis pipeline.}
    \label{fig:pipe}
\end{figure}

In this study, we used a pipeline architecture to allow for multimodal analysis, as shown in Figure \ref{fig:pipe}. As previously mentioned, we collected three distinct modalities: biometric, facial, and behavioral data. Several feature extraction methods and pre-processing steps can be done based on the modality type and the task that we are dealing with. Possible suggestions are provided in the following sections:

\subsubsection{Feature Extraction}
Feature extraction plays an important role in drowsiness modeling by adding meaningful information to raw data. The features that can be derived from different modalities are explained below:

\begin{itemize}
    \item \textbf{Biometric Signals:} For signals like HR, IBI, BVP, EDA, TEMP, SpO2, and pulse rate, we can extract several time-domain features, such as the mean, standard deviation, and root mean square. These features help us understand the body’s normal state and how it changes over time \cite{kreibig2010autonomic}. We can also extract frequency-domain features, such as power spectral density from HRV and frequency analysis of EDA signals \cite{zhang2018effects} to help identify phasic (rapid) and tonic (baseline) components of arousal changes \cite{bodaghi2024multimodal}. Additionally, data like ACC and motion sensors offer insight into physical activity levels. Features like mean acceleration magnitude, entropy, and movement frequency are useful for detecting reduced movement. Other statistical metrics, such as skewness, kurtosis, and signal entropy, help capture small and large changes over time \cite{hosseini2023multimodal}.
 
    \item \textbf{Facial Data:} Facial video data provides both visual and depth-based information for identifying signs of drowsiness. From these videos, FLs and FAUs are the most common features that can be extracted using DLib \cite{king2009dlib} or MediaPipe \cite{lugaresi2019mediapipe}. The key points are those related to the eyes, mouth, and eyebrows. Commonly used features include the Eye Aspect Ratio (EAR), Mouth Aspect Ratio (MAR), blink frequency, mouth opening angle, and the percentage of eye closure over time (PERCLOS). These metrics are well-known indicators of drowsiness \cite{bodaghi2024adaptive}. We also track head movements, such as nodding or tilting. Depth maps help us to extract spatio-temporal features, such as frame-to-frame changes in landmark positions. These can help capture tiny movements of the eyelid or small changes in posture, which are often early signs of fatigue.

    \item \textbf{Behavioral Data:} Behavioral data captures the physical movement and interaction of the driver with the vehicle, which is affected by drowsiness. Potential features include extracting PLs from video using tools like OpenPose\cite{cao2021openpose} and MediaPipe. Features, such as posture angles, movement amplitude, and stillness duration, can be extracted with the focus on key points like shoulders, spine, and head. For grip pressure and telemetry data, time-domain, frequency-domain, and statistical features are mostly useful. For example, average grip pressure, grip duration, speed variability, and engine RPM patterns can provide valuable information about the driver's state \cite{bajaj2023system}.
    
\end{itemize}

\subsubsection{Data Pre-processing}
We do pre-processing to make sure data from different modalities are cleaned, aligned and transformed into suitable formats for fusion and classification. In the public version of the dataset, each modality has a unique sampling frequency, as explained in Section \ref{sec9}. Here are some suggested pre-processing steps for each modality:

\begin{itemize}
    \item \textbf{Biometric Signals:} These signals can easily become noisy from sudden movements, sensor placement, or environmental factors like temperature and humidity. To clean them, techniques like band-pass filtering are suitable to remove high-frequency noise \cite{pourbemany2023survey}. At the same time, a method called independent component analysis (ICA) can help separate motion artifacts from signals like HR, BVP, EDA, and ACC data \cite{zhu2024emotion}. To make sure everything is consistent across different subjects, we can normalize the data for each participant separately. Also, resampling or interpolation techniques can address different sampling rate issues so they line up properly with each other.

    \item \textbf{Facial Data:} Facial video data require pre-processing to address differences in lighting, occlusions, and camera-specific noise. Standardizing video frames to a fixed resolution facilitates consistent feature extraction. Histogram equalization\cite{zhang2023self} or adaptive gamma correction\cite{yang2021low} can mitigate lighting variations, while median or Gaussian filters help remove depth and pixel-level noise \cite{li2023extended}. Additionally, optical flow-based interpolation can reconstruct missing frames to maintain temporal consistency in video sequences.

    \item \textbf{Behavioral Data:} Behavioral data, including video and time-series data, require similar pre-processing techniques as those mentioned for biometric and facial data. Filtering techniques are useful for reducing sensor or environmental noise, while cropping regions of interest ensure that only relevant areas are analyzed. Normalization and resampling methods standardize data across participants and align different sampling rates. Additionally, outlier removal techniques help eliminate anomalies.

    \item \textbf{Multimodal:} Before integrating modalities into a multimodal framework, additional pre-processing is needed for synchronization, alignment, and compatibility. Temporal alignment means adjusting all modalities to a single sampling rate, such as 4 Hz, to make synchronized data fusion easier. windowing methods can be used to segment the data into fixed-length time windows for analysis. Normalization can be helpful in consistency across modalities and prevents bias due to scale differences. Finally, methods like Principal Component Analysis (PCA) and manifold learning can be used to decrease the data dimension, which reduces the computing cost, particularly when working with data that has many features\cite{bodaghi2024multimodal}.

\end{itemize}

\subsubsection{Integration Across Modalities}
\begin{figure}[htbp]
    \centering
    \begin{subfigure}[b]{0.26\textwidth}
        \centering
        \includegraphics[width=\linewidth]{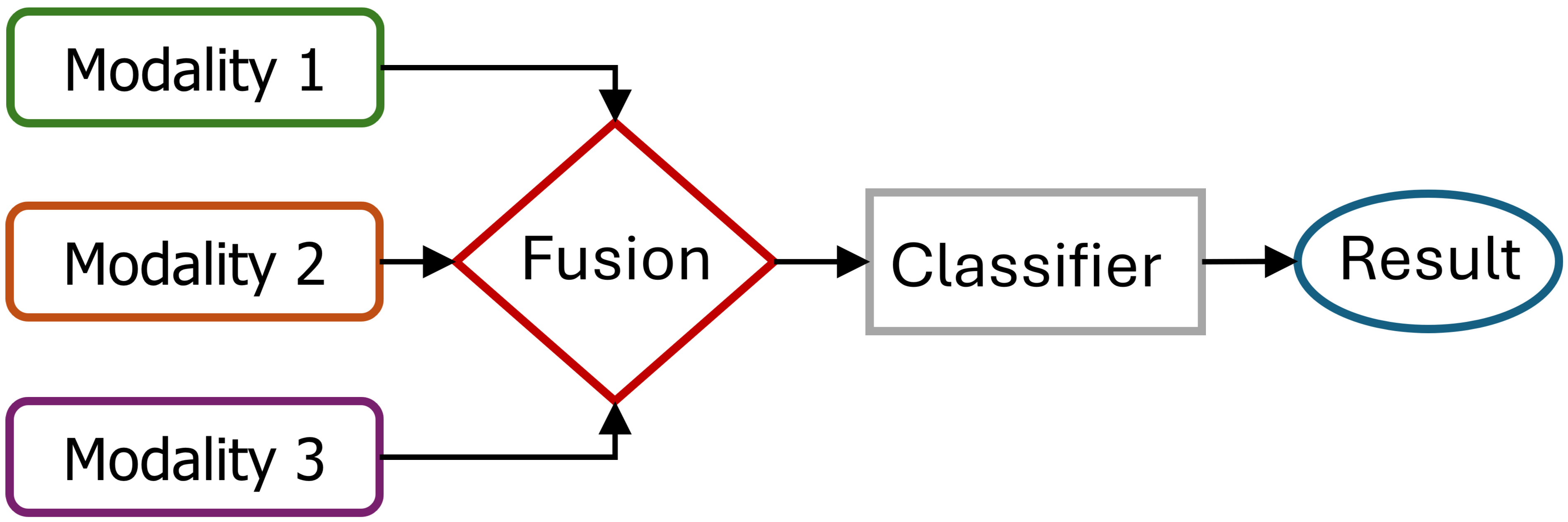}
        \caption{Early-fusion.}
        \label{fig:early}
    \end{subfigure}
    \hspace{1mm}
    \begin{subfigure}[b]{0.26\textwidth}
        \centering
        \includegraphics[width=\linewidth]{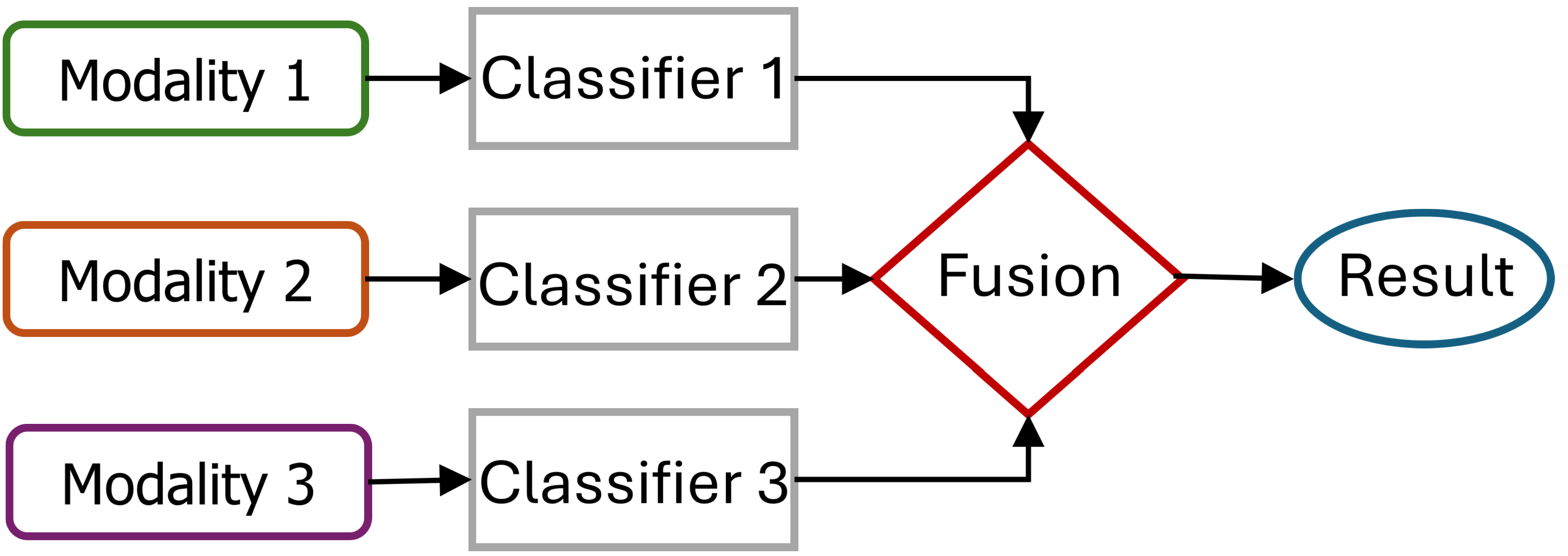}
        \caption{Late-fusion.}
        \label{fig:late}
    \end{subfigure}
    \hspace{1mm}
    \begin{subfigure}[b]{0.42\textwidth}
        \centering
        \includegraphics[width=\linewidth]{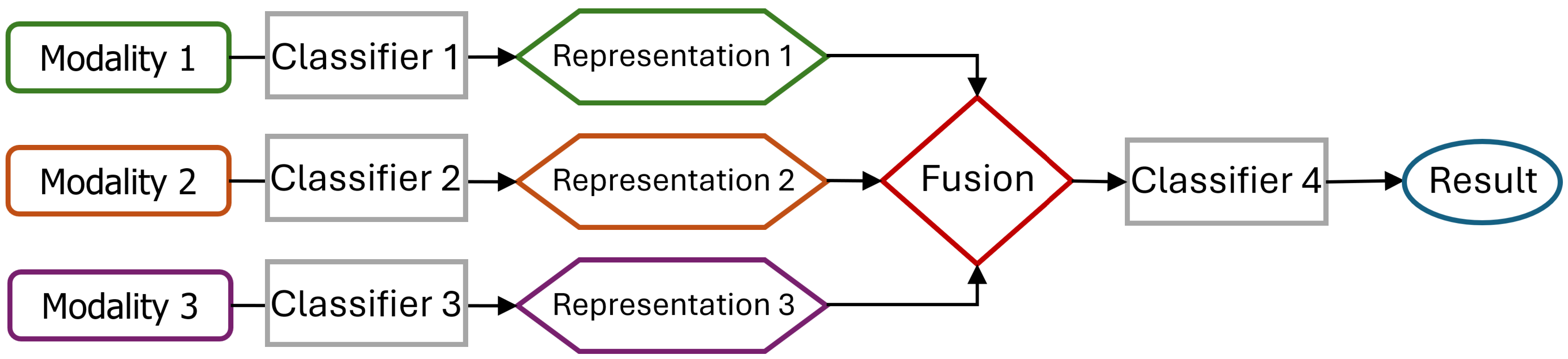}
        \caption{Intermediate-fusion.}
        \label{fig:inter}
    \end{subfigure}
    
    \caption{Conventional methods for multimodal data fusion.}
    \label{fig:fusions}
\end{figure}

After pre-processing, several fusion strategies can be applied depending on the task:

• \textbf{Early-Fusion:} This technique merges raw or basic features from all modalities into one feature vector before performing classification. It helps identify connections between modalities, but it can lead to challenges with higher data dimensions and noise (see Figure \ref{fig:early}). 

• \textbf{Late-Fusion:} In this method, each modality is processed independently to provide separate classification, and then these individual decisions are combined using techniques like majority voting, basic averaging, or weighted combination. This method is stronger against noise for each modality, and it is more flexible in data pre-processing, but it might miss relationships between modalities (see Figure \ref{fig:late}).

• \textbf{Intermediate-Fusion:} This method combines modalities at an intermediate stage after extracting higher-level features and representations but before the final classification. This allows a balance for capturing correlations between modalities while maintaining flexibility in feature representation. It has a higher computational cost and better performance in most cases compared to early and late fusion approaches (see Figure \ref{fig:inter}).

\section{Technical validation}\label{sec10}
In this section, we provide a quality validation for the various captured signals to determine their suitability for driver drowsiness detection and to examine the correlation between the signals and different drowsiness levels. To achieve this, we applied statistical tests to assess the significance of the signals in relation to drowsiness levels, considering both direct and inverse proportionality, as well as individual variability. We used signal-to-noise ratio (SNR) analysis to evaluate signal quality. Label consistency was verified using Cohen’s Kappa for inter-rater agreement. We have also tested the multimodal aspect of the data by performing classification tasks using two well-known machine learning models with k-fold cross-validation.

\subsection{Physiological Validation}
\begin{figure}[htbp]
    \centering
    \begin{subfigure}[b]{0.48\textwidth}
        \centering
        \includegraphics[width=\linewidth]{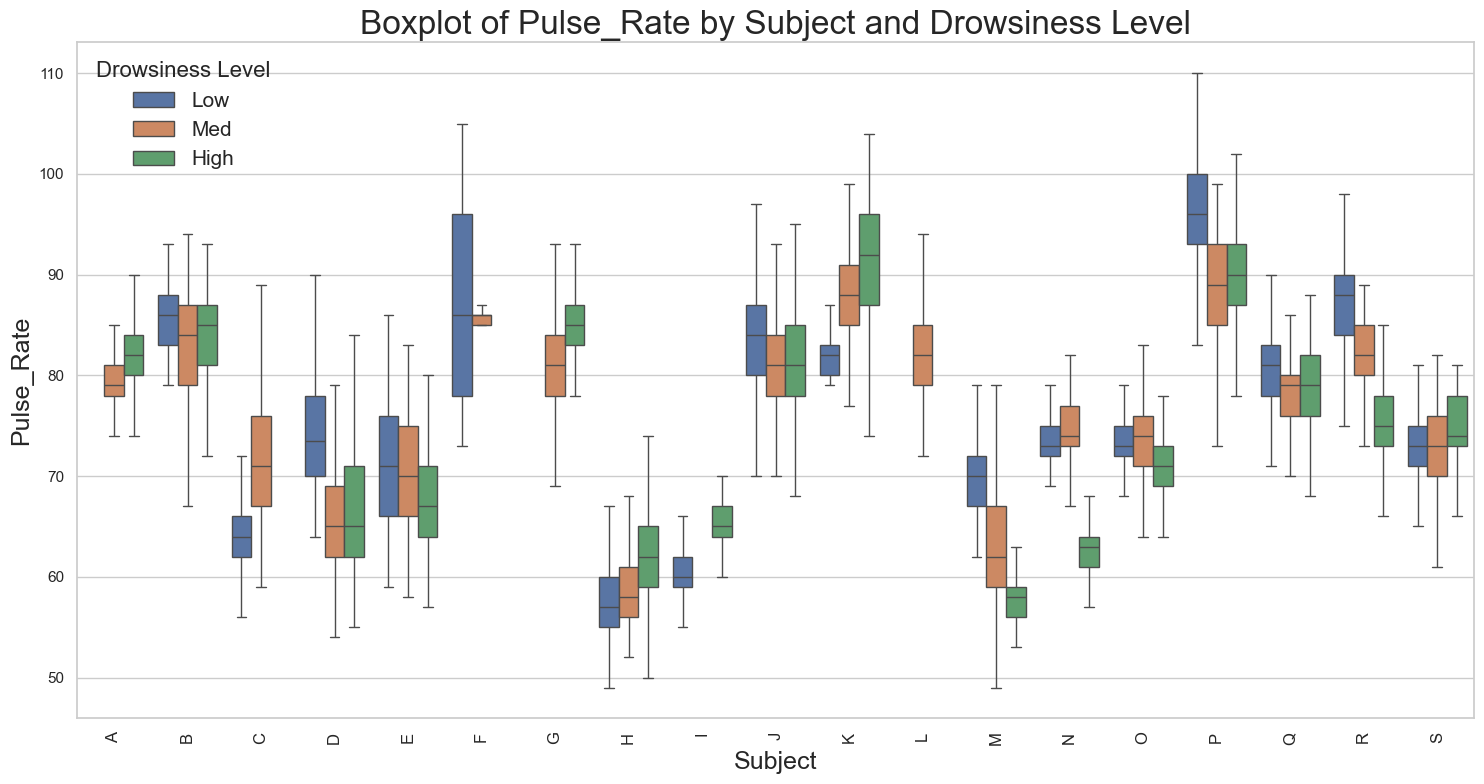}
        \caption{Pulse Rate (PR).}
        \label{fig:PR}
    \end{subfigure}
    \hspace{1mm} 
    \begin{subfigure}[b]{0.48\textwidth}
        \centering
        \includegraphics[width=\linewidth]{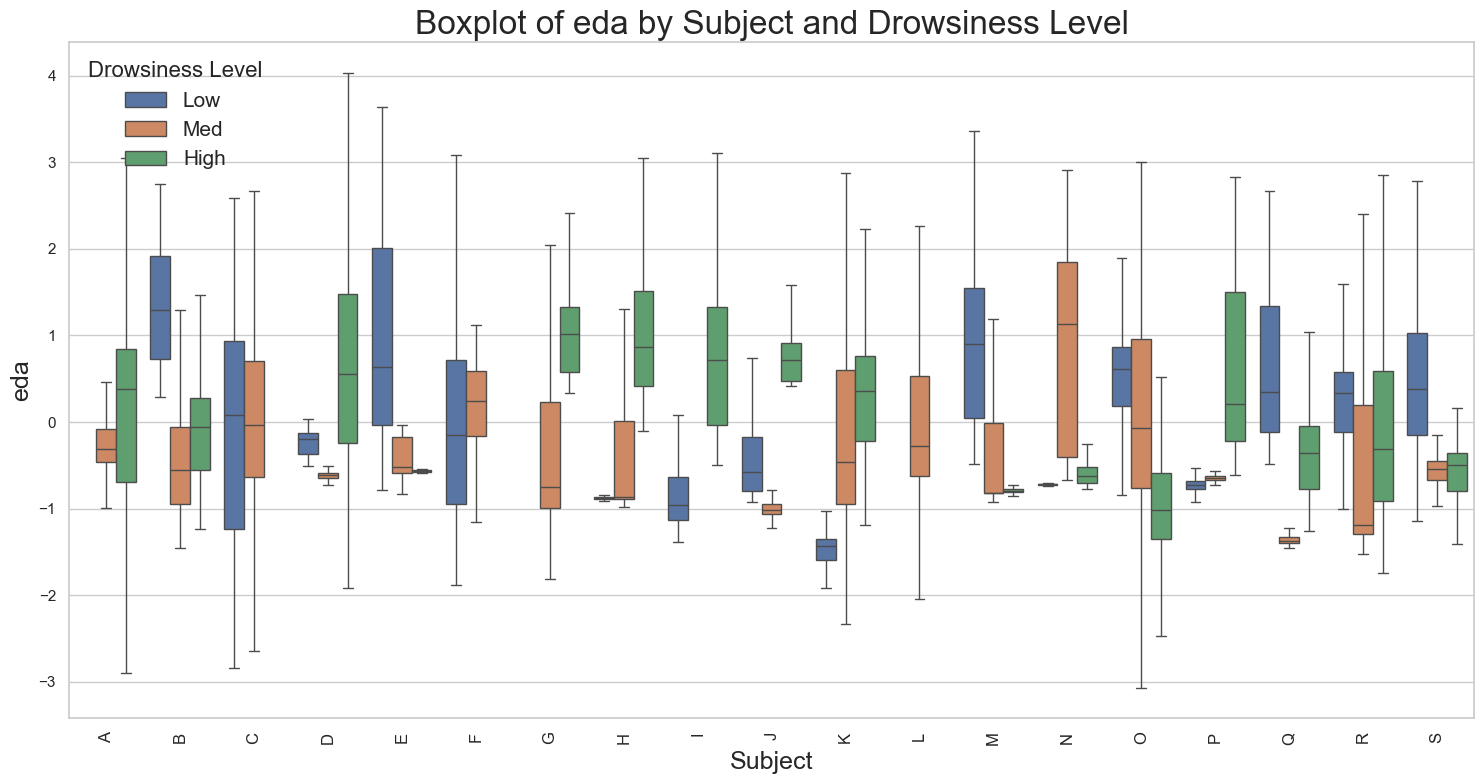}
        \caption{Electrodermal activity (EDA).}
        \label{fig:EDA}
    \end{subfigure}
    
    \vspace{0.5mm} 
    \begin{subfigure}[b]{0.48\textwidth}
        \centering
        \includegraphics[width=\linewidth]{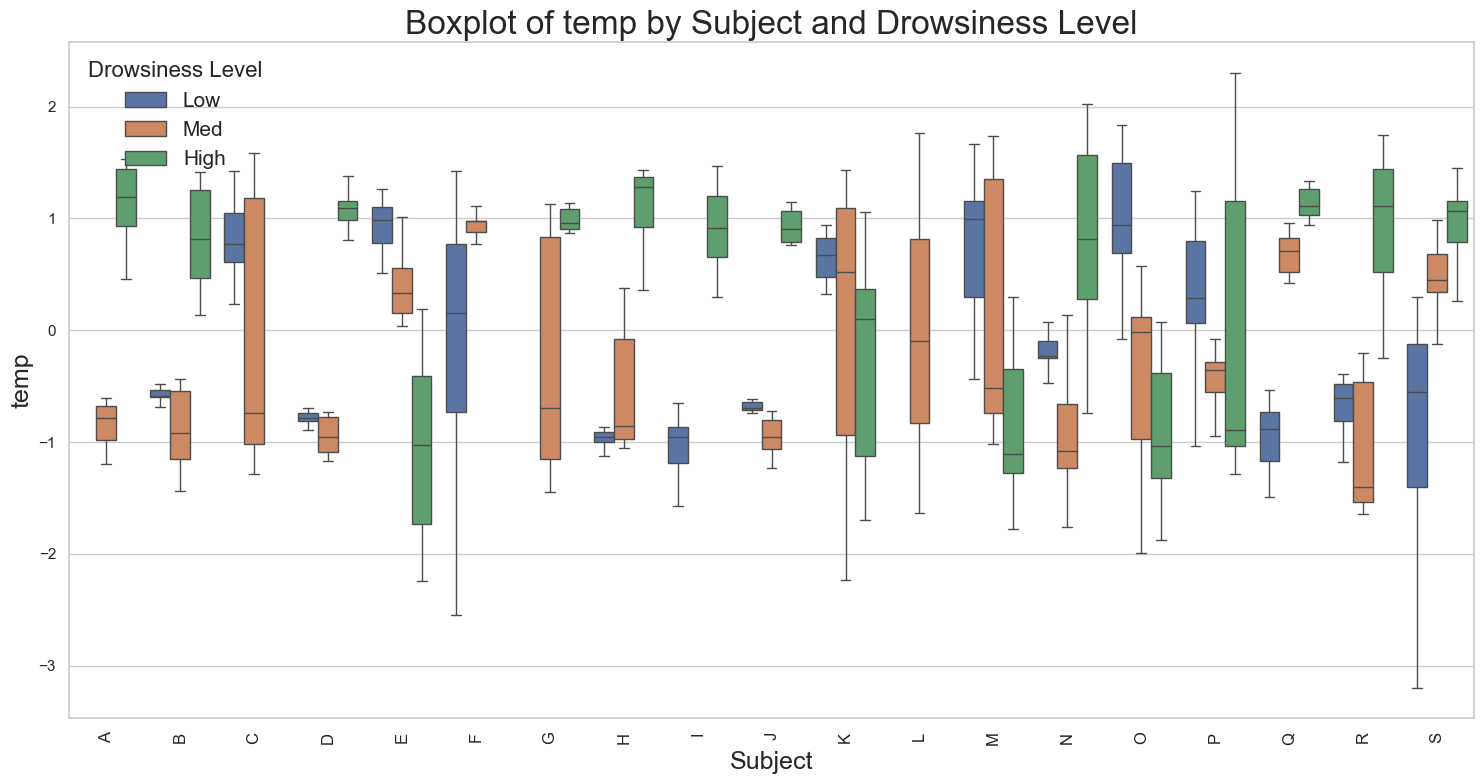}
        \caption{Skin temperature (TEMP).}
        \label{fig:temp}
    \end{subfigure}
    \hspace{1mm} 
    \begin{subfigure}[b]{0.48\textwidth}
        \centering
        \includegraphics[width=\linewidth]{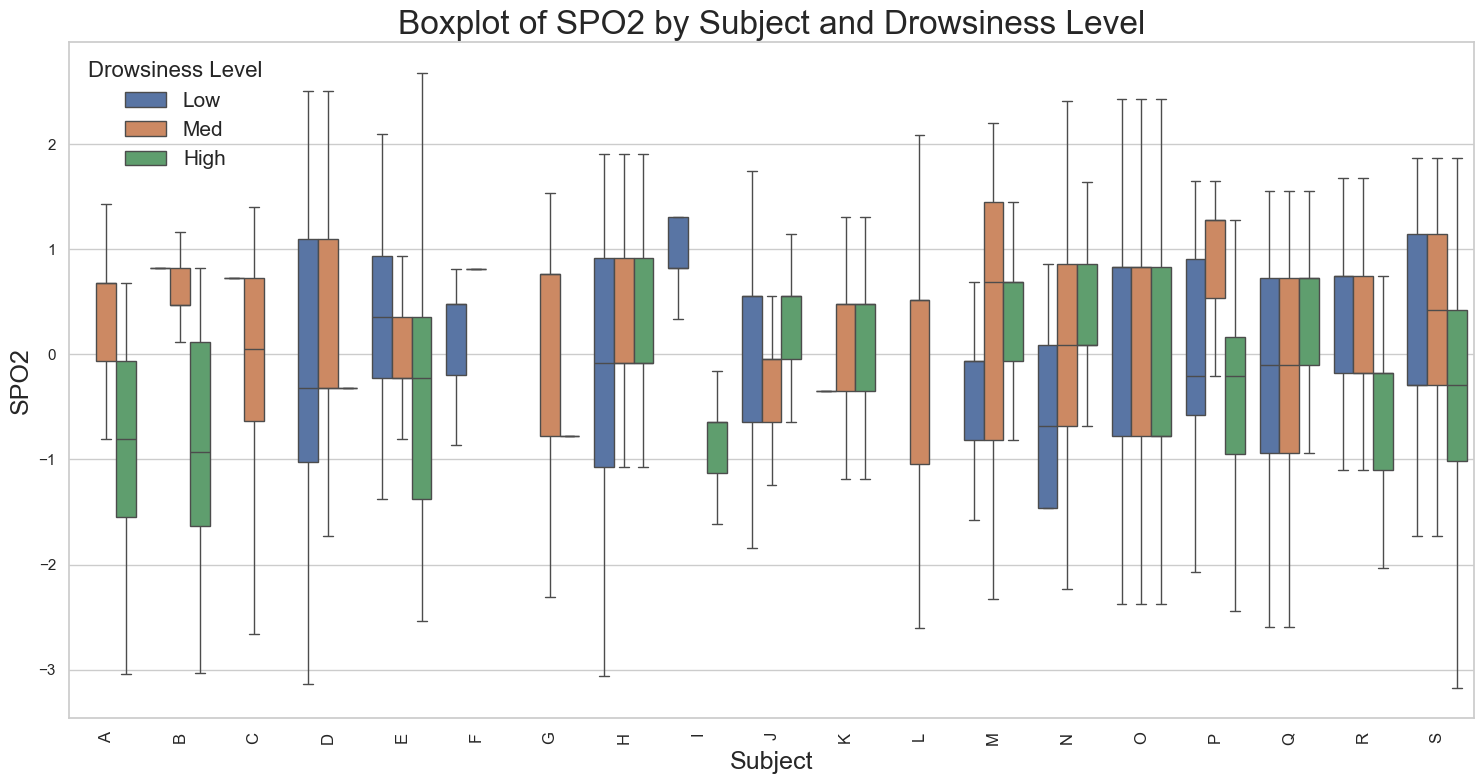}
        \caption{Blood oxygen saturation (SPO2).}
        \label{fig:spo2}
    \end{subfigure}

    \caption{Box plots of biometric signals for all the subjects and three levels of drowsiness.}
    \label{fig:box}
\end{figure}

To validate the collected biometric signals and how they change with different levels of drowsiness, we present box plots (Figure \ref{fig:box}) that show the distribution of each signal across all participants. These plots are divided into three categories of drowsiness: low, medium, and high. For illustration, we removed the outliers for each signal per subject and then standardized them to have a mean of zero and a standard deviation of one to be consistent for all subjects. These distributions help validate the reliability of the collected signals and highlight patterns of change in response to increasing drowsiness levels. These plots show the inherent inter-subject variability in physiological data that is necessary for understanding individual differences in response to drowsiness.


\subsubsection{Statistical Analysis}
To validate the physiological data, we first tested whether it followed a normal distribution or not, using the Shapiro-Wilk\cite{royston1992approximating} test. The results indicated that the data was not normally distributed. Because of this, we used the Kruskal-Wallis Test \cite{mckight2010kruskal}, which is a non-parametric option instead of ANOVA. This test looks at the rankings of different groups to see if the differences in how the signals are spread across drowsiness levels are meaningful or just random chance. It assesses whether the samples come from the same distribution or exhibit fundamental differences. Since the  number of drowsiness level labels is limited (KSS=9), we categorized drowsiness levels into Low, Medium, and High, as outlined below:
\begin{itemize} \item High: Drowsiness levels greater than 6
\item Medium: Drowsiness levels between 4 and 6 
\item Low: Drowsiness levels less than 4 \end{itemize}

 All tests show significant differences between 'Low and 'Medium' drowsiness levels with P-values less than 0.001, so that we can reject the null hypothesis, except for BVP with a p-value of 0.059. Similarly, all measurements exhibit significant differences between 'Medium' and 'High' drowsiness levels except ACC\_X, with a p-value of 0.098. Finally, all measurements show significant differences between 'Low' and 'High' drowsiness levels for all biometric signals except BVP, with the p-value of 0.055.

We also assessed the statistical significance of the data using the Linear mixed effects model\cite{pinheiro2006mixed} that extends traditional linear regression by incorporating both fixed and random effects. Fixed effects capture the overall population-level relationships between independent variables and the dependent variable, while random effects account for variations at the group or individual level. This dual structure allows for more accurate data modeling with hierarchical or clustered arrangements, such as repeated measurements from the same subjects or data grouped by categories. We used the same binning for the drowsiness level for linear models with mixed effects. We explored the relationship between various physiological parameters and levels of drowsiness by incorporating inter-subject variability among 19 participants. The linear model with mixed effects shows that the BVP signals do not correlate significantly with drowsiness, so we can not reject the null hypothesis. The remaining biometric signals showed a significant correlation (P$<$0.001), with one exception in HR when comparing medium and high drowsiness levels with a p-value of 0.2. As we used two different sensors for capturing heart-related activity, PR shows more consistent results across drowsiness levels, so we recommend relying on PR data for analysis.

\begin{table}[ht]
\centering
\caption{Mixed-Effects Model Results for Biometric Signals.}

\begin{tabular}{|c|c|c|c|c|c|c|c|}
\hline
\textbf{Signal}      & \textbf{Coef.} & \textbf{Est.} & \textbf{Std.Err} & \textbf{Z-score} & \textbf{P-value} & \textbf{Group Var} & \textbf{Residual Std} \\ \hline
\textbf{EDA}         & Intercept            & 0.565             & 0.196           & 2.888            & 0.004            & 0.727              & 0.510                 \\ \hline
                     & T.Low                & 0.220             & 0.002            & 90.490            & \textless 0.001  &                    &                       \\ \hline
                     & T.Med                & 0.075             & 0.002            & 33.452           & \textless 0.001  &                    &                       \\ \hline
\textbf{TEMP}        & Intercept            & 33.099            & 0.264            & 125.508          & \textless 0.001  & 1.321              & 1.085                 \\ \hline
\textbf{}            & T.Low                & -1.120            & 0.005            & -216.406         & \textless 0.001  &                    &                       \\ \hline
\textbf{}            & T.Med                & -1.397            & 0.005            & -293.733         & \textless 0.001  &                    &                       \\ \hline
\textbf{HR}         & Intercept            & 74.303            & 1.688            & 44.025          & \textless 0.001  & 54.113             & 6.567                 \\ \hline
\textbf{}            & T.Low                & 1.983             & 0.031            & 63.328          & \textless 0.001  &                    &                       \\ \hline
                     & T.Med                & -0.037            & 0.029            & -1.272          & 0.2  &                    &                       \\ \hline
\textbf{BVP}         & Intercept            & 0.037             & 0.669            & 0.055            & 0.956            & 8.043              & 51.520               \\ \hline
                     & T.Low                & -0.025            & 0.245            & -0.101           & 0.920            &                    &                       \\ \hline
                     & T.Med                & -0.030            & 0.225            & -0.132           & 0.895            &                    &                       \\ \hline
\textbf{SPO2}        & Intercept            & 94.946            & 0.255            & 372.445         & \textless 0.001  & 1.234              & 1.674                 \\ \hline
                     & T.Low                & 0.745             & 0.008            & 93.340          & \textless 0.001  &                    &                       \\ \hline
                     & T.Med                & 0.851             & 0.007            & 116.014          & \textless 0.001  &                    &                       \\ \hline
\textbf{PR} & Intercept            & 74.981            & 1.954            & 38.366           & \textless 0.001  & 72.564              & 5.827                \\ \hline
                     & T.Low                & 2.733             & 0.028            & 98.388           & \textless 0.001  &                    &                       \\ \hline
                     & T.Med                & 0.758            & 0.026            & 29.698           & \textless 0.001  &                    &                       \\ \hline
\end{tabular}

\label{tab:mixedeffects}
\end{table}

Based on the test results presented in Table \ref{tab:mixedeffects}, the following observation was made:
\begin{itemize}
    \item \textbf{EDA} decreased with increased drowsiness levels, indicating heightened sympathetic nervous system activity when individuals are more drowsy. A group variance of 0.727 suggests small variability in EDA levels across subjects. The z-score values indicate highly significant effects, meaning these coefficients differ significantly from zero. The residual standard deviation indicates a small amount of unexplained variability in the data, but the model's fit remains reasonably accurate.
    \item \textbf{TEMP} increased with increasing drowsiness levels, reflecting a growth in skin temperature as individuals become more drowsy, as the body tends to warm up in Peripheral areas \cite{krauchi2001circadian}. This occurs due to increased blood flow during the transition to sleep, particularly in the fingers and hands, where wearables typically measure temperature\cite{tai2023association, szymusiak2018body}. A group variance of 1.321 indicates moderate variability in temperature across subjects. The z-scores are extremely large and significantly different from zero. The residual standard deviation suggests that while the model fits the data well, there remains a moderate amount of unexplained variability in the temperature measurements.
    
    \item \textbf{SPO2} exhibited slight increases at lower drowsiness levels, which may reflect improved respiratory or cardiovascular function associated with alertness. A small group variance indicates small variability across subjects. The exceptionally large z-scores and their high significance underline the robustness of these associations.
    
    \item \textbf{PR} increased with lower drowsiness levels. This suggests heightened cardiovascular activity when individuals are more alert. The large group variance indicates substantial variability in pulse rate across subjects, reflecting individual differences in heart rate responses. The residual standard deviation points to moderate unexplained variability in pulse rate, suggesting that other factors beyond drowsiness levels may also influence these measurements.

\end{itemize}

\subsection{Signal Quality Assessment}\label{sec11}
To make sure that our dataset is reliable, we evaluated signal quality for biometric signals, behavioral data, and video data. Firstly, for biometric signals (HR, EDA, TEMP, SpO2, pulse rate, ACC) and behavioral data (grip pressure), we conduct a Signal-to-noise ratio (SNR) analysis using a 5-minute rest period. Specifically, for each subject and session, we computed the SNR of the signals using a third-order Butterworth low-pass filter to separate the signal from noise. Then, the SNR was calculated as $\text{SNR} = 10 \log_{10} \left( \frac{P_{\text{signal}}}{P_{\text{noise}}} \right)$, where $P_{\text{signal}}$ and $P_{\text{noise}}$ are the power of the filtered signal and noise, respectively \cite{yang2024multimodal}. The noise was calculated by subtracting the filtered signal from the original. For HR, SPO2, and pulse rate, a cutoff frequency of 1.5 Hz was used to preserve cardiac-related components\cite{patterson2009flexible}. EDA was filtered at 1.5 Hz to retain both tonic and phasic components\cite{posada2020innovations}, and TEMP and ACC were filtered at 1.0 Hz due to their slower-varying nature. The grip pressure was also set to 1.0 Hz to minimize high-frequency fluctuations \cite{day2021low}. The results are presented in table \ref{tab:snr}:

\begin{table}[htbp]
\caption{SNR analysis results for biometric and behavioral signals.}
\begin{tabular}{|c|c|c|c|c|}
\hline
\textbf{Signal}        & \textbf{Avg SNR (dB)} & \textbf{SD (dB)} & \textbf{Min-Max (dB)} & \textbf{\% \textless 10 dB} \\ \hline
\textbf{HR}            & 45.67                 & 3.19             & 38.48 - 51.53         & 0                           \\ \hline
\textbf{EDA}           & 21.20                 & 10.35            & 1.55 - 43.26          & 14.29                       \\ \hline
\textbf{TEMP}          & 26.44                 & 5.03             & 15.56 - 36.13         & 0                           \\ \hline
\textbf{SPO2}          & 30.23                 & 3.83             & 23.12 - 37.93         & 0                           \\ \hline
\textbf{Pulse Rate}    & 31.95                 & 3.91             & 25.96 - 40.53         & 0                           \\ \hline
\textbf{ACC}           & 19.59                 & 5.11             & 7.10 - 27.08          & 8.57                        \\ \hline
\textbf{Grip Pressure} & 14.68                 & 5.45             & 5.34 - 31.14          & 20                          \\ \hline
\end{tabular}
\label{tab:snr}
\end{table}

Overall, the SNR results confirm that most signals were captured with high quality. HR showed the highest signal quality, and grip pressure showed the lowest SNR, probably due to sensor contact quality. 

In addition, we also assessed the quality of the recorded video data. To verify the integrity of the video recordings, we computed the frame drop rate by comparing the number of frames extracted using OpenCV (VideoCapture function) with the expected number of frames based on the video’s frame rate and duration. This showed a frame drop rate of 0.00\%, confirming that the videos were recorded without loss and are suitable for further behavioral or facial expression analysis.

\subsection{Inter-Rater Reliability}
To check the consistency of the labels, we asked an expert to watch some of the videos (30\% of the total data) and label them independently based on KSS criteria. Then we compared participants' self-reported labels with expert annotations using Cohen's Kappa measures of inter-rater agreement. For the 9-level KSS scale, the unweighted Kappa was $\kappa = 0.619$, indicating substantial agreement. To consider the ordinal nature of the rating and penalize larger disagreements more heavily, we used quadratic-weighted Kappa, which resulted in $\kappa = 0.967$ (95\%  CI: [0.951, 0.978]), showing almost perfect agreement. For the binned 3-level, we computed the unweighted Kappa. resulting in a $\kappa = 0.88$, also reflecting high consistency.

\subsection{Validation of Multimodal Data}\label{sec12}
To assess the quality of the UL-DD dataset, we performed a classification task as a form of validation. For each data type and signal, we began by combining the raw synchronized data from all subjects and sessions. To reduce computational cost and maintain consistency across all modalities, we set the sample rate frequency to 4 Hz for all data types. Missing values were handled separately for each subject and session using linear interpolation with forward fill. After interpolation, the data were normalized using MinMaxScaler. We performed three-level drowsiness classification, labeling the data as Low ($KSS<4$), Medium ($4<= KSS <=6$), and High ($KSS>6$). Since our focus was on validating the dataset, we did not apply any additional pre-processing steps. To understand the value of each modality, we performed both unimodal and multimodal classification. 

We used two commonly applied machine learning models aligned with \cite{yang2024multimodal}: The Support Vector Machine (SVM) with an RBF kernel (C=0.1, gamma='auto',class\_weight='balanced') and the Random Forest (RF) classifier (max\_depth=5, n\_estimators=100, min\_samples\_leaf=10). To evaluate model performance, we used 5-fold cross-validation (k=5) to ensure generalizability. Folds were stratified to maintain the distribution of drowsiness levels across subjects, accounting for class imbalance and inter-subject variability.

\begin{figure}[htbp]
    \centering
    \includegraphics[width=0.9\textwidth]{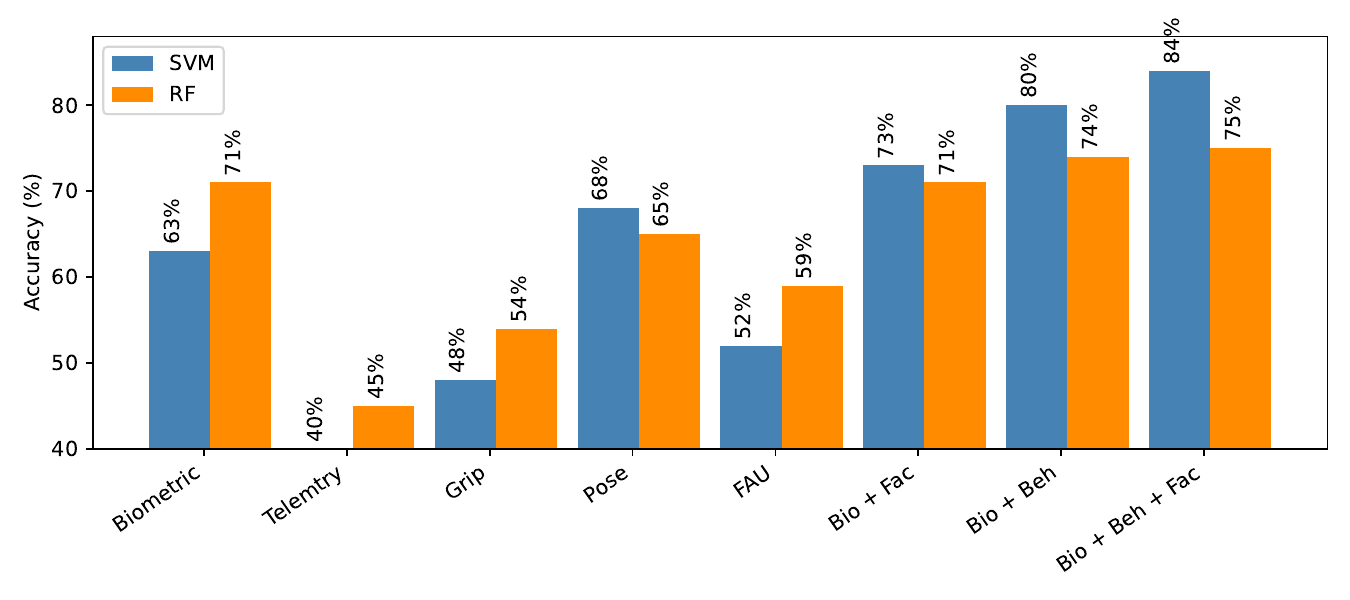}
    \caption{Comparing the performance of SVM and RF models for different data types and combinations.}
    \label{fig:result}
\end{figure}

Figure \ref{fig:result} presents the classification accuracy results for individual modalities and multimodal combinations of modalities. For the multimodal task, we employed an early fusion approach, where features from different data types are integrated before feeding them into the model. In this way, we are evaluating not only the individual modality contribution, but also how well they work together. Biometric signals served as the baseline modality for this comparison.
Interestingly, the performance trend was consistent across both models. For unimodal data, the RF model performed slightly better, specifically for biometric signals, grip pressure, telemetry, and FAUs. On the other hand, SVM showed stronger results when multiple data types were combined. The most effective combination was biometric signals, behavioral data, and facial features, demonstrating the strength of multimodal learning and the dataset’s validity for drowsiness detection.

\section{Usage notes}\label{sec14}
The original version of the UL-DD dataset can be accessed upon request. The data is organized by participants and session, stored in CSV and MP4 formats for ease of use. For pre-processing, we recommend the mentioned steps in Section \ref{pre}. The dataset can be used to develop and test multimodal drowsiness detection models, analyze the relationships between facial, behavioral, and biometric indicators of drowsiness.

\section{Code availability}\label{sec15}
Readers can access all the code and tutorials along with the dataset. This includes scripts for combining raw data of all subjects, adjusting sample frequency, assembling datasets by combining different modalities, and labeling. Additionally, the scripts for both unimodal and multimodal analysis are available in the repository. We also provided scripts for the biometric technical analysis and feature extraction code for video data, including 2D/3D FLs, FAUs, and PLs. A detailed README file is included for more information and guidance.

\section{Limitation and future work}\label{sec16}
The UL-DD dataset offers a comprehensive and versatile multimodal resource for detecting early signs of drowsiness, yet it comes with several limitations. The first constraint that comes to attention is the number of participants with unbalanced gender population that limits the diversity and may not be generalizable for the broader population. Additionally, using a driving simulator in a controlled lab environment is different from real-world conditions that come with different lighting, weather, traffic, and road conditions. The technical constraints include different signal qualities across modalities and the absence of video data for some subjects, which affects the usage of data for some multimodal analyses, even though extracted features were provided to address this gap. There were recruitment challenges due to the inability to offer incentives, which reduced the participant rate and gender balance.

For future work, we will address these limitations by expanding the number of participants and improving diversity. Collecting real-world data beyond 40 minutes to include different environmental conditions, such as different lighting and occlusion, would enhance the validity. Adding more sensors like seat pressure to capture weight distribution or advanced devices for capturing brain and eye activity could enrich the multimodal aspect of the dataset. The publicity of the dataset will provide opportunities for future research to build robust and generalizable drowsiness detection models by leveraging its diverse signals. These advancements could boost UL-DD’s contributions to driver monitoring systems and road safety research.

\backmatter

\section{Acknowledgments}

This work has been partially funded by NSF (CVDI 10a.005.UL\_TAU) and TietoEvry.

\section{Competing Interests}
The authors declare no competing interests.

\bibliography{sn-bibliography}

\end{document}